\documentclass[10pt,twocolumn,letterpaper]{article}

\usepackage{cvpr}              % To produce the CAMERA-READY version
\usepackage[dvipsnames]{xcolor}

\usepackage{amsmath,amssymb,amsfonts}
\usepackage{alltt}
\usepackage{accents}
\usepackage{booktabs}
\usepackage{bm}
\usepackage{bbold}
\usepackage{color}
\usepackage{colortbl}
\usepackage{caption}
\usepackage{enumerate}
\usepackage{epstopdf}
\usepackage{esvect}
\usepackage{gensymb}
\usepackage{adjustbox}

\usepackage{mathtools}
\usepackage{multirow}
\usepackage{mwe}
\usepackage{pifont}
\usepackage{picinpar}
\usepackage{pbox}

\usepackage{soul}
\usepackage{siunitx}
\usepackage{subcaption} %Dont' use duplicated with subfigure package
\usepackage{stmaryrd}

\usepackage{textcomp}
\usepackage{url}
\usepackage{algpseudocode}
\usepackage{algorithm}
\usepackage{dsfont}

\algrenewcommand\algorithmicrequire{\textbf{Input:}}
\algrenewcommand\algorithmicensure{\textbf{Output:}}
\makeatletter % changes the catcode of @ to 11
\newcommand{\vE}{\vec{\pmb{E}}\@ifnextchar{^}{\,}{}}
\makeatother % changes the catcode of @ back to 12

\makeatletter
\newcommand{\thickhline}{
    \noalign {\ifnum 0=`}\fi \hrule height 1pt
    \futurelet \reserved@a \@xhline
}
\makeatother

\newcommand{\comment}[1]{}
\newcommand{\cmark}{\ding{51}}%
\newcommand{\xmark}{\ding{55}}%

\definecolor{cvprblue}{rgb}{0.21,0.49,0.74}
\usepackage[pagebackref,breaklinks,colorlinks,citecolor=cvprblue]{hyperref}

\def\paperID{14769} % *** Enter the Paper ID here
\def\confName{CVPR}
\def\confYear{2024}

\title{Image-Object-Specific Prompt Learning for \\Few-Shot Class-Incremental Learning}

\author{In-Ug Yoon\\
KAIST \\
{\tt\small iuyoon@rit.kaist.ac.kr}
\and
Tae-Min Choi\\
KAIST\\
{\tt\small tmchoi@rit.kaist.ac.kr}
\and
Sun-Kyung Lee\\
KAIST\\
{\tt\small sklee@rit.kaist.ac.kr}
\and
Young-Min Kim\\
Samsung Research\\
{\tt\small ym1012.kim@samsung.com}
\and
Jong-Hwan Kim\\
KAIST\\
{\tt\small jhkim@rit.kaist.ac.kr}
}

\begin{document}
\maketitle
\begin{abstract}
While numerous studies in few-shot class-incremental learning (FSCIL) have been conducted, attaining satisfactory performance, particularly during incremental sessions, remains a significant challenge. A key issue is that encoders, trained on extensive base session datasets, tend to underperform in subsequent incremental sessions. In this context, the Contrastive Language-Image Pre-training (CLIP) model, known for its ability to generalize to unseen classes, emerges as a promising option for FSCIL tasks. 
%However, developing a learning scheme to effectively utilize CLIP for FSCIL can be challenging, primarily due to the issues such as  forgetting of the knowledge for the past sessions and the capability to learn the new session knowledges.
%, which may potentially lead to final session scores that are lower than those achieved by zero-shot learning.
%However, tailoring a learning scheme to utilize CLIP for FSCIL effectively can be difficult, leading to forgetting or overfitting during incremental session training, and potentially resulting in even lower final session scores compared to zero-shot learning.
%However, developing a learning scheme that effectively utilizes CLIP for FSCIL can be challenging, even showing lower score at the final session compared to zero-shot learning. This is due to issues such as forgetting previous knowledge and insufficient learning ability for new sessions. To address the issues, we introduce a novel training framework centered around developing image-object-specific (IOS) classifiers. Here, an IOS classifier is designed to focus on particular attributes (e.g., wings or wheels) of class objects. Our framework is designed from an FSCIL perspective to maintain previously acquired knowledge and quickly adapt to new sessions without forgetting or overfitting.
However, developing a learning scheme that effectively utilizes CLIP for FSCIL can be challenging, even showing a lower score in the final session than zero-shot learning.
This is due to issues such as forgetting previous knowledge and insufficient learning ability for new sessions.
We introduce a novel training framework centered around developing image-object-specific (IOS) classifiers to address the issues. Here, an IOS classifier is designed to focus on class objects' particular attributes (e.g., wings or wheels). Our framework is designed from an FSCIL perspective to maintain previously acquired knowledge and quickly adapt to new sessions without forgetting or overfitting.
Our method consistently outperforms existing state-of-the-art techniques across datasets such as miniImageNet, CIFAR100, and CUB200. Also, we conduct further experiments to confirm the efficacy of our model in generating IOS classifiers and undertake ablation studies to evaluate the contribution of each component within our architecture.

\end{abstract}    
%%%%%%%%% BODY TEXT
\section{Introduction}

Class-Incremental Learning (CIL) is a deep learning algorithmic scheme designed to adapt quickly and efficiently to changes in the environment, while maintaining high performance. CIL is composed of sequential tasks, and for each task, only the training data from the current task should be accessible, to account for issues related to memory storage and data security.
However, conventional CIL presupposes a substantial amount of training data for incremental sessions \cite{yan2021dynamically, li2023multi, tang2023prompt, khan2023introducing}, which can be challenging and time-consuming to obtain. Furthermore, the acquisition of such a dataset can impede the ability to swiftly adapt to changes. By comparison, humans can quickly adjust to new environments with only a limited number of training examples, without forgetting previous knowledge.
Therefore, it is only logical to expect a deep learning algorithm that can adapt to new environments using just a few samples, without forgetting previous knowledge. This is precisely the goal of Few-Shot Class-Incremental Learning (FSCIL).

\begin{figure}
\centering
\includegraphics[width=0.75\linewidth]{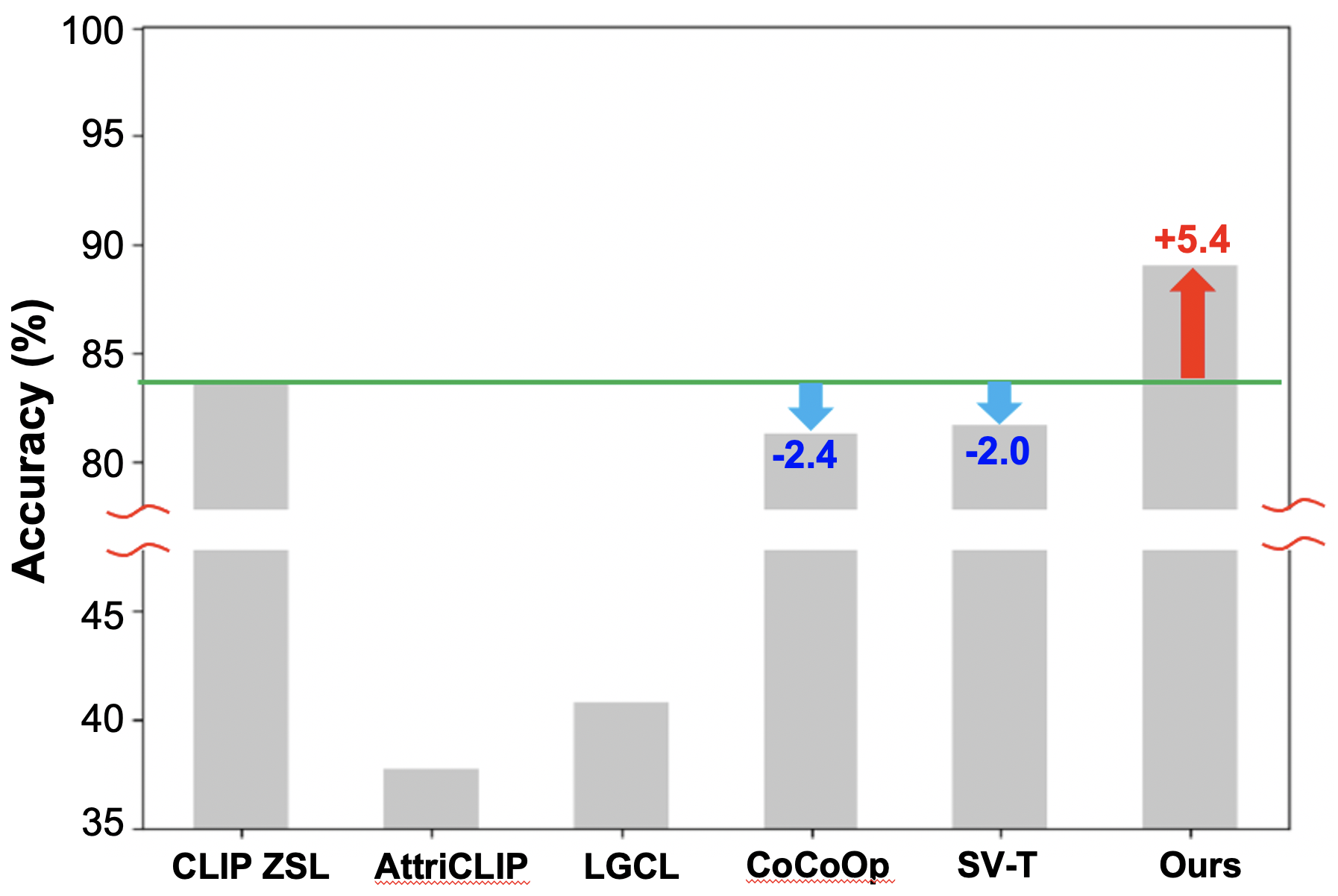}
\caption{
Comparison of various methods employing the CLIP model in terms of classification performance on all seen classes at the final session under the FSCIL setting.
The results from CLIP zero-shot learning (ZSL) are remarkably impressive.
However, other methods resulted in performance even lower than the CLIP ZSL outcomes.
%, likely due to the forgetting and insufficient learning of new tasks during the incremental session training.
More detailed information on this experiment can be found in our supplementary materials.
%Classification performance at the final session with the FSCIL setting utilizing the CLIP model. CLIP zero-shot learning results are notably impressive. However, diverse approaches to further training with the CLIP model led to even worse performance than the zero-shot learning performance, likely due to issues related to forgetting and overfitting. Further experimental details are described in our supplementary materials.
%Classification performance at the final session with the FSCIL setting utilizing CLIP model.% with ViT-B backbone network. CLIP zero-shot learning results are notably impressive. 
%We compare diverse methods utilizing CLIP model, class-incremental learning
%However, diverse approaches to further training with CLIP model led to even worse performance than the zero-shot learning performance, likely due to issues related to forgetting and overfitting. Further experimental details are described in our supplementary materials.
%However, while diverse approaches to further training with CLIP model initially led to enhanced performance in earlier sessions, this trend subsequently reversed in later sessions, likely due to issues related to forgetting and overfitting.
}
\label{fig:intro_compare_figure}
\vspace{-3mm}
\end{figure}

%Maintaining the datasets from past environments not only poses a memory storage issue but also raises data security concerns.
%This feature poses the main challenge of continual learning: severe forgetting of knowledge from past environments while overfitting to the current environment.
%Several attempts, including knowledge distillation, feature regularization, utilizing exemplar sets, and knowledge replay, have been proposed to address this issue. TBDxxx refs.
%However, typical continual learning assumes the availability of a sufficient amount of training dataset for new environments, which can be difficult and time-consuming to acquire. Furthermore, acquiring such a dataset hinders the ability to quickly adapt to changes.

\begin{figure*}[t]
    \centering 
    \begin{subfigure}{6.5cm}
    \includegraphics[width=6.5cm]{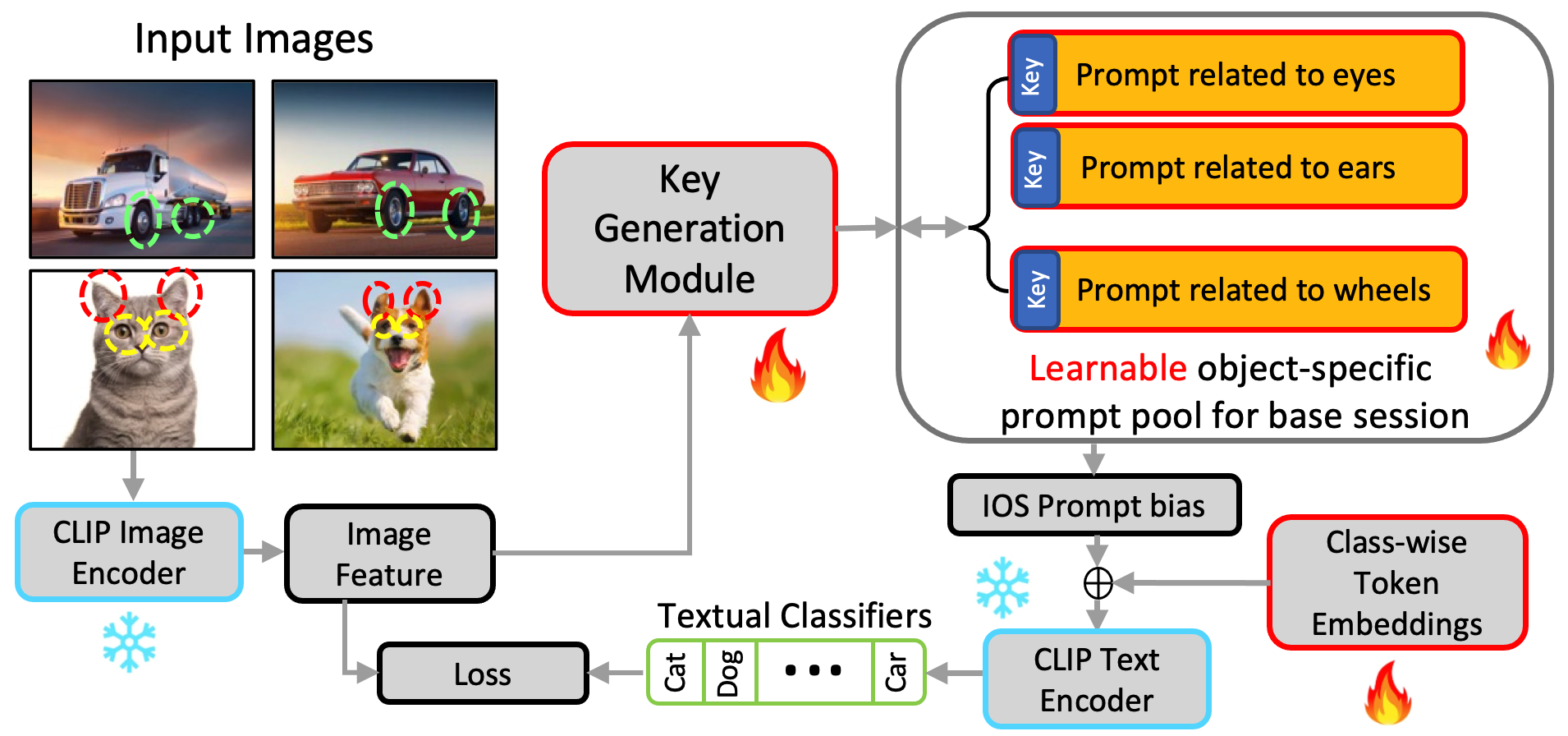}
    \caption{Base session}
    \label{fig:loss_direction_a}
    \end{subfigure}
    \quad
    \begin{subfigure}{8.0cm}
    \includegraphics[width=8.0cm]{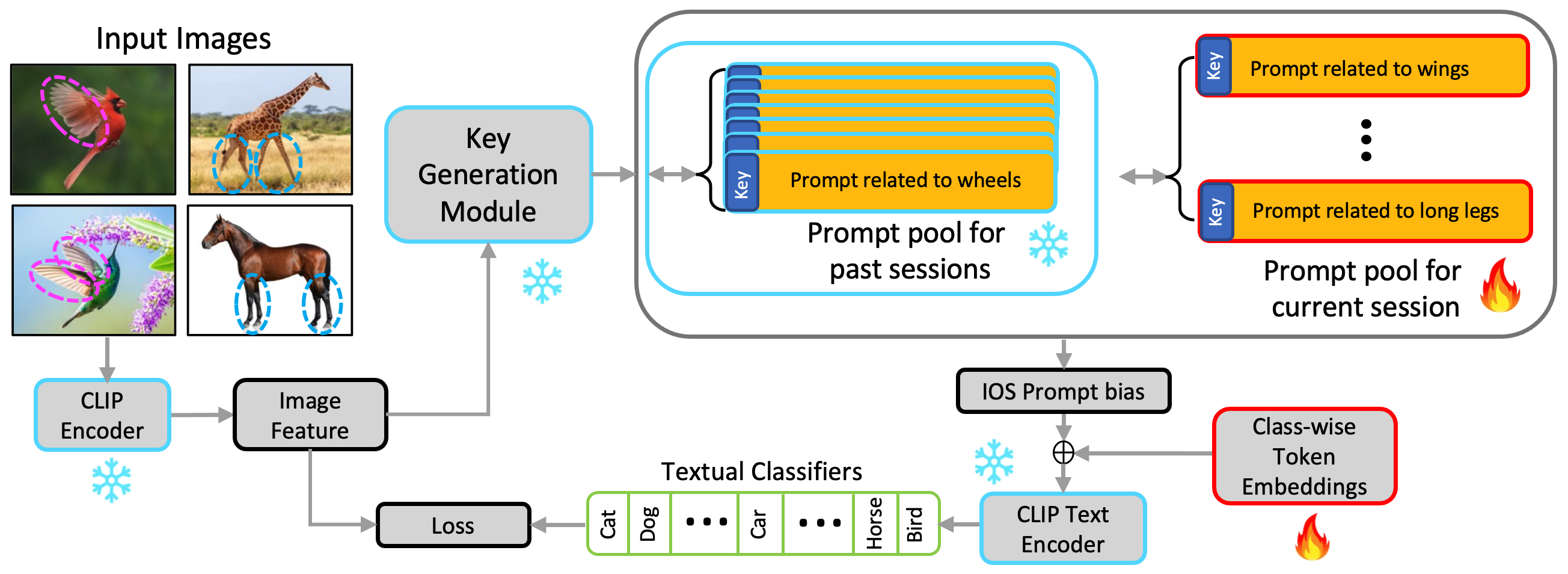}
    \caption{Incremental sessions}
    \label{fig:loss_direction_b}
    \end{subfigure}
    \caption{
    Conceptual illustration of our proposed training architecture for FSCIL. Our central idea revolves around generating and utilizing textual classifiers that encode the image-object-specific characteristics, harnessed using key-prompt pairs for each session. Icons with a fire symbol indicate learnable modules, while those with a freeze symbol represent frozen parameters.
    %In the base session, using a sufficient base session dataset, we train the network to generate IOS prompt bias by providing more attention to the object characteristics instead of the background of the image. The goal of training this model is twofold: firstly, to improve the base session classification performance, and secondly, to capture object features of incremental sessions quickly and accurately by focusing on object regions. In the incremental sessions, the key generation module and feature prompts for past sessions are frozen to prevent forgetting, and only feature prompts for the current session are learnable.
    }
    \label{fig:concept}
\vspace{-3mm}%
\end{figure*}

%In contrast, humans can fast adapt to the new environments only with limited amount of training examples, without forgetting past knowledges.
%So it is natural to demand deep learning algorithm which can adapt to new environment only with few samples without forgetting the past knowledges, which is aim of few-shot class-incremental learning (FSCIL).
%FSCIL is consisted of base session and incremental sessions. Sufficient amount of training set for the classes in base session are provided while only limited number of training set is given for classes in incremental sessions.

%In contrast, humans can quickly adapt to new environments with only a limited number of training examples, without forgetting past knowledge. It is therefore natural to demand a deep learning algorithm that can adapt to new environments with only a few samples, without forgetting past knowledge, which is the aim of few-shot class-incremental learning (FSCIL).
FSCIL comprises base and incremental sessions. An ample amount of training data is provided in the base session, while the classes in the incremental sessions are given only a limited amount of training data.
The primary challenge of FSCIL, similar to CIL, is to minimize forgetting of past sessions and prevent overfitting to the current session during the incremental learning process.
Various strategies have been employed to tackle FSCIL, including the use of knowledge distillation \cite{dong2021few,cheraghian2021semantic}, pre-training of the feature extractor \cite{chen2020incremental}, and knowledge replay \cite{liu2022few}.

%However, previous FSCIL methods had clear limitations on the performance, especially for the incremental session classes. The reason is as follows. Since fine-tuning the massive image encoder with a small number of training data in incremental sessions led to severe forgetting, previous approaches froze the model or added simple networks for training. Consequently, the performance relied on how well the pre-trained image encoder can extract features for new classes. However, image encoders have been trained on the base training dataset, resulting in significantly lower performance for new classes than the base session classes TBDxxx ref. Thus, a pre-trained model with high generalizability, extracting effective features from both seen and unseen class images, is necessitated for achieving high FSCIL task performance.
However, previous FSCIL methods have displayed clear limitations in their performance, particularly with regards to the classes in the incremental sessions. The underlying reason is as follows: Fine-tuning the substantial image encoder with a small quantity of training data in incremental sessions led to severe forgetting. In response, prior approaches either froze the model or incorporated simple networks for training \cite{zhang2021few, zhou2022forward}. As a result, performance became largely dependent on the efficacy of the pre-trained image encoder in extracting features for new classes. However, image encoders have been trained on the base training dataset, which leads to significantly lower performance for new classes as compared to base session classes \cite{cheraghian2021semantic, tao2020few, zhang2021few}.
Therefore, a pre-trained model that exhibits high generalizability and is capable of extracting effective features from both seen and unseen class images, is needed to achieve superior performance in FSCIL tasks.

Recent research leveraging Contrastive Language-Image Pre-trained (CLIP) models has gained significant momentum, mainly due to their versatility, anchored in their paired image and text encoders. These models have also demonstrated their broad applicability, achieving notable results even in zero-shot classification tasks \cite{radford2021learning}.
In Fig. \ref{fig:intro_compare_figure}, we examine the FSCIL performance of various CLIP-based methodologies. This includes applications in CIL (e.g., AttriCLIP \cite{wang2023attriclip} and LGCL \cite{khan2023introducing}), generalization to unseen classes (e.g., CoCoOp \cite{zhou2022conditional}), and approaches for FSCIL (e.g., SV-T \cite{qiu2023semantic}).
Our observations reveal that methods involving additional training exhibit a significant decline in performance as sessions progress, with final session outcomes occasionally falling below the zero-shot learning capabilities of CLIP. This decline is likely attributable to severe forgetting of past session knowledge and insufficient learning for the new session during incremental session training. Moreover, strategies intended to mitigate these issues during incremental session training \cite{wang2023attriclip, khan2023introducing}, assuming ample training data for these sessions, have shown limited effectiveness in learning new tasks in FSCIL settings.

In this paper, addressing these challenges, we aim to develop an novel approach for FSCIL that capitalizes on the strengths of the CLIP model while circumventing the problems of forgetting and overfitting.
We specifically focus on designing the learning structure to leverage image-object-specific (IOS) features and classifiers. The aspects of the learning structure, as illustrated in Fig. \ref{fig:concept}, are as follows:
%In this paper, we focus on designing the learning structure to leverage the CLIP model for the FSCIL setting, focusing particularly on image-object-specific (IOS) features and classifiers. The features of the learning structure, as illustrated in Fig. \ref{fig:concept}, are as follows:
%we aim to learn and utilize image-object-specific prompt for the FSCIL.

%First, we proposed an image-object-specific (IOS) prompt generation module. Image features obtained by the CLIP image encoder have a formidable amount of information, including class object-related, background-related, etc. Therefore, we designed to train the IOS prompt generation module using a sufficient base session training set to improve the base session classification performance and capture object feature prompts of the incremental sessions quickly and accurately by focusing on the object-related regions.
Firstly, we propose a learning structure designed to capture the inherent IOS features from the image and incorporate them to the generated classifiers, which we term as IOS classifiers.
%the learning structure to capture the IOS characteristics from the image and reflect 
Image features obtained by the CLIP image encoder have a formidable amount of information, including class object-related (which we refer as IOS), background-related, and more. 
However, for the classification, IOS characteristics (e.g., wings or wheels) plays the key role.
Thus, we expected that focusing on the IOS characteristics could enhance both accuracy and efficiency for the classification.
%Thus, we design the learning structure to extract the IOS features and reflect to the generated classifiers.
%Thus, with a ample training set within the base session, we train the network to extract IOS features from the image. These features are then utilized for generating IOS classifiers.
%Thus, with ample training set within the base session, we train the network to extract IOS features, which is used for generating IOS classifiers, from the image feature.

Secondly, our learning scheme is designed to enhance the capability to acquire knowledge in incremental sessions with limited training data. Previously learned IOS characteristics may facilitate learning new classes that share similar attributes. Additionally, to support the acquisition of new IOS characteristics for the new session, we propose an initialization strategy to address the convergence issues that occur when proceeding with random initialization with only a few training data.
Finally, we tailored the learning scheme to minimize forgetting and overfitting during the incremental sessions. Specifically, we have carefully designed the update targets for the training phase of these sessions to prevent overfitting to the current session and forgetting past knowledge.
%we froze the parts of the network that worked for the previous sessions, which included the key map and the key-prompt pairs from the past. Only the key-prompt pair and class-wise token embeddings for the current session were updated.
In short, our contributions are as follows:

%Finally, we propose some tricks for fast convergence during the incremental session. Since the number of training data for each incremental session is very low and may be insufficient for converging parameters from the random initialization. Thus we introduce some tricks to utilize a limited number of training data and fast converge parameters belonging to incremental sessions.
%Finally, we propose an initialization strategy for rapid convergence of the IOS-related modules. Shortage of training data in incremental sessions may lead to convergence issues when using random initialization. Thus we design an additional technique utilizing the class information inherent in the class names, which could assist the convergence of IOS-related modules which aims to encode class object characteristics. In short, our contributions are as follows:
\begin{itemize}
\item{We propose an image-object-specific prompt generation module to utilize object-related information within images when generating textual classifiers.}
%\item{Our learning scheme is specifically designed to enhance the capability of learning new tasks. This enhancement is achieved through modules focused on capturing Image-Object-Specific (IOS) characteristics and a proposed initialization strategy aimed at resolving convergence issues during incremental sessions.}
\item{Our learning scheme is designed to enhance the capability of learning new tasks, achieved via modules that capture IOS characteristics and a proposed initialization strategy to address convergence issues.}
\item{We propose an optimized learning strategy to utilize the IOS classifiers for FSCIL, with perspective of minimizing the forgetting and overfitting issues during the incremental sessions.}
%\item{We propose an optimized learning strategy to utilize the IOS classifiers for FSCIL, with perspective of minimizing the forgetting and overfitting issues during the incremental sessions.}
%\item{We propose a training structure to leverage CLIP model for FSCIL task, preventing forgetting to the past and overfitting to the current session.}
%\item{We propose an initialization technique for IOS-related modules for quick convergence during incremental sessions with limited training data.}
%\item{We further analyze the effects of each module based on diverse ablation studies.}
\end{itemize}

\section{Related Work}

\textbf{Few-shot class-incremental learning.} %이거 최근 논문들 다수 추가. 
The most challenging aspect of FSCIL is assimilating new knowledge for novel tasks without causing the forgetting of previously acquired knowledge. 
First line of approaches attempted to minimize the forgetting.
The primary research objective in this area is to devise strategies to tackle this problem of forgetfulness. 

One proposed solution entails employing neural gas to maintain the feature topology between the classes of the base and new sessions \cite{tao2020few}. Another method involves data replay \cite{liu2022few}, where a certain amount of prior task data is stored in exemplar sets and utilized for replay during incremental sessions.
Also, the technique of knowledge distillation has been applied, utilizing the model from earlier sessions as a guiding teacher model to mitigate forgetting \cite{dong2021few}. 
Other approaches focus on the pre-training method during the base session to prepare for the upcoming incremental sessions. Meta-learning approaches aim to enhance the network's adaptability to new tasks by creating episodic scenarios using base session data \cite{hersche2022constrained, zhou2022few}. Strategies that preserve regions of the embedding space for upcoming classes have also demonstrated performance improvements \cite{zhou2022forward}.
From the classifier's perspective, an attention model was trained in the base session to balance the classifiers between past and new sessions during the incremental sessions \cite{zhang2021few}. However, despite these diverse attempts, the overall performance generally fell short of expectations, particularly for the incremental sessions.

\noindent\textbf{Prompt Learning on CLIP Models for Continual Learning.}
Multi-modal large-scale models such as CLIP models recently showed remarkable usability, including the model generalizability for unseen classes, as shown in the zero-shot classification performance.
To preserve the model expandability, researches added and updated the learnable prompts.
Initial researches split dataset into base and new classes to evaluate.
One line of work attempted to add prompts within the image and text encoders to reflect the learned knowledges \cite{miao2023mudpt}.
Also replacing the basic text prompt to learnable prompts enhanced image classification performance \cite{zhou2022learning}.
%-xxxTBD et al CoOp CoCoOp 
Moreover, adding an image-specific prompt bias, prompt formed from image feature, to the learnable text prompts to form the textual classifiers enhanced the model performance on both seen and unseen classes \cite{zhou2022conditional}.
Focusing on extracting object-specific information from the image to form the prompt bias also showed significant performance \cite{goswami2023contextual}.

There were also approaches to utilize the CLIP model for class-incremental tasks.
Learning task-wise attribute prompts for each task were proceeded, encoding both object-related and background-related attributes  to form textual classifiers \cite{wang2023attriclip}.
Learning both task-level and class-level guidance prompt for the sequential tasks verified the effectiveness \cite{khan2023introducing}.
Also aggregated structure of the image and textual features using the attention model was designed to form textual classifiers \cite{zhou2023learning}.

However, these methods were designed to train in incremental sessions based on a large number of current session training datasets and exemplar sets. Consequently, research has been carried out to utilize the CLIP model in FSCIL settings, which only have a few training data for incremental sessions. 
Such works have either added an average pooling and MLP layer to the pre-trained image transformer \cite{qiu2023semantic}, or adapted learnable parameters to both the image and text encoders for each task \cite{d2023multimodal}.

\begin{figure*}[t]
    \centering 
    \begin{subfigure}{16cm}
    \includegraphics[width=14.5cm]{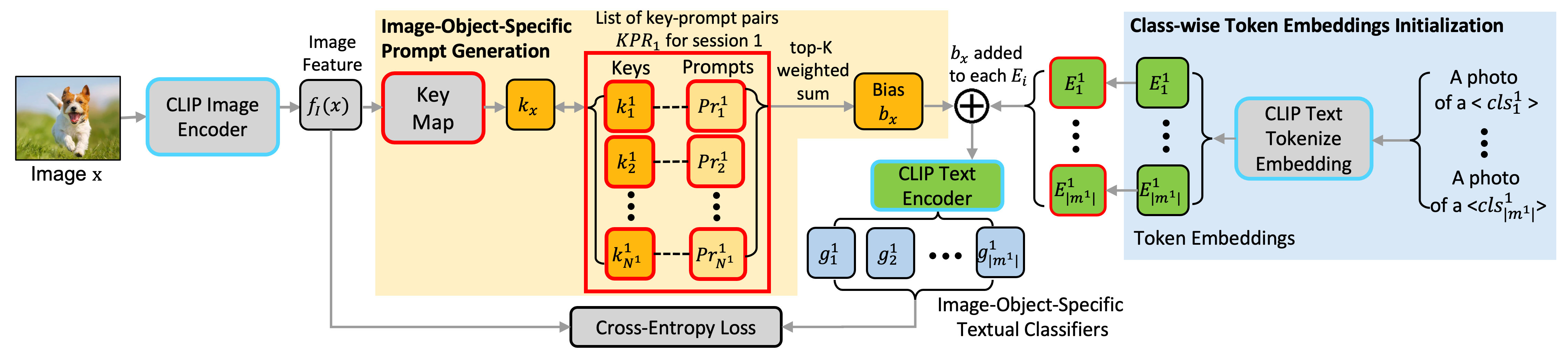}    
    \caption{Base session ($t=1$)}
    \label{fig:detail_learning_schematics_base}
    \end{subfigure}
    \begin{subfigure}{16cm}
    \includegraphics[width=14.5cm]{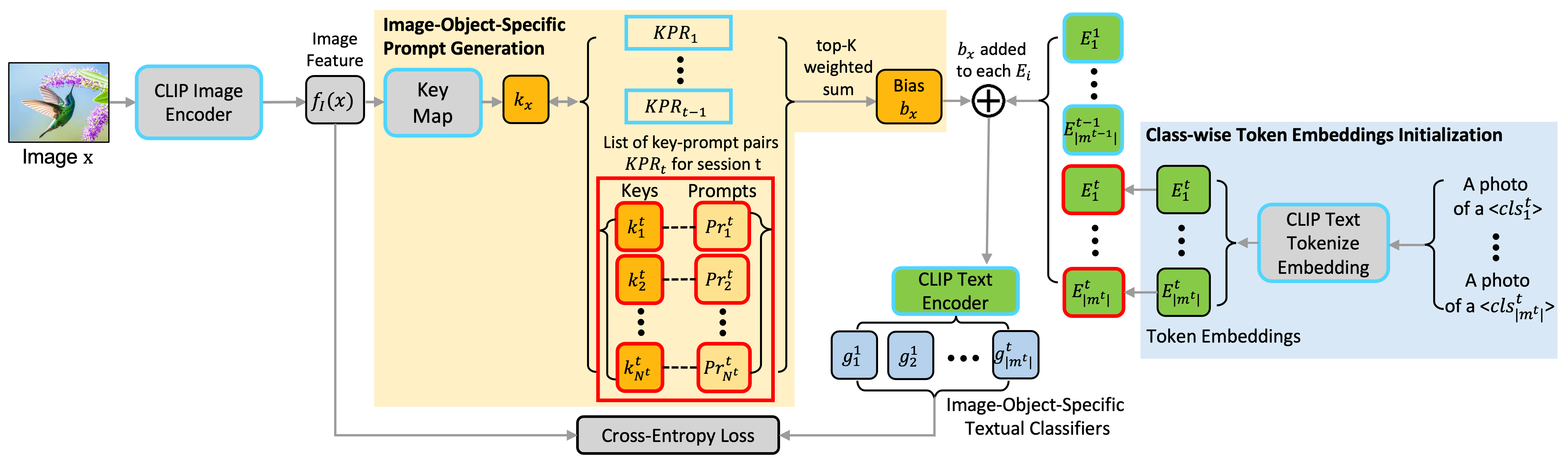}
    \caption{Incremental session ($t>1$)}
    \end{subfigure}
    \caption{
    An overview of our training framework encompasses the ‘class-wise token embedding initialization' and ‘image-object-specific prompt generation' modules. We leverage these modules to capture the IOS attributes within images and reflect them to the generated classifiers. Note that we elaborately designed the updatability of each module in every session to mitigate forgetting, where a red borderline module indicates learnable while a blue borderline module indicates frozen. 
    %(a) In the base session, first, we initialize the class-wise token embeddings $(E^t_i)$ which is learnable throughout the training. The purpose of using class-wise embeddings is to enhance the model capacity to capture the features of each class. Then the image-object-specific prompt generation module generates prompt bias from the image feature  by matching the generated key $k_x$ with the list of key-prompt pairs $(k^t_i, Pr^t_i)$ and combining the highly related prompts. Finally, we combine the prompt bias and the class-wise token embeddings to generate textual classifiers with image-object-specific feature encoded. The cross-entorpy loss between the image feature and the generated classifiers is calculated for the model update. Note that red borderline means the learnable and the blue borderline means frozen. (b) In the incremental session, most of the training procedure is similar to the base session except the updated parameters. The parameters containing the knowledge for the past sessions is frozen and only the parameters for the new session is updated.
    }
    \label{fig:detailed_schematics}
\vspace{-2mm}%
\end{figure*}

\section{Problem Set-up}

\label{sec:problem_formulation}
FSCIL consists of sequential tasks having datasets $\mathcal{D}=\{\mathcal{D}^1,\mathcal{D}^2,...,\mathcal{D}^T\}$,
where $\mathcal{D}^t=\{\mathcal{D}^t_{tr},\mathcal{D}^t_{te}\}$ denotes training and test dataset for session \textit{t}.
Training set of session \textit{t}, $D^t_{tr}$ consists of class labels $\mathcal{Y}^t_{tr}=\mathcal{Y}^t=\{y^t_1,...,y^t_{m^t}\}$, where $m^t$ is the number of classes in session $t$.
Note that different sessions have no overlapped classes, $\textit{i.e.}$ $\forall i,j$ and $i\neq j$, $\mathcal{Y}^i\cap\mathcal{Y}^j=\varnothing$.
Test set of session \textit{t}, $D^t_{te}$ is composed of class labels $\mathcal{Y}^1\cup\mathcal{Y}^2\cdot\cdot\cdot\cup\:\mathcal{Y}^t$.
For convenience, we denote $\mathcal{Y}^{i:j}=\mathcal{Y}^i\cup\cdot\cdot\cdot\cup\:\mathcal{Y}^j$ for $i<j$.
For $\textit{N}$-way $\textit{K}$-shot FSCIL setting, $|\mathcal{D}^t_{tr}|=K\times N$, $m^t=N$ for $t>1$. 
%For example, in the popular benchmark dataset CUB200, there are 100 classes in the base session and 100 classes for incremental sessions. For 10-way 5-shot setting, the number of sessions $T=11$, $K=5$ and $|\mathcal{C}^t|=10$ for $t>1$.
%each class has 500 training images, while in each incremental session, only 5 classes are available for training and each class only has 5 images. 
%FSCIL defines a harsh problem setting, where the severe data imbalance and scarcity
%problems will further exacerbate knowledge forgetting in incremental learning.

CLIP image encoder and text encoder are denoted as $f_{I}(\cdot)\in\mathbb{R}^D$ and $f_{T}(\cdot)\in\mathbb{R}^D$, where $D$ is the embedded dimension.
With the text input $x_{text}$ and length function $L(x_{text})$ which outputs the length of the tokenized input text, the CLIP tokenized embedding function is denoted as $h(x_{text})\in\mathbb{R}^{L(x_{text})\times D}$.

%In this paper, we employ a coreset to utilize representative image features for classes of past sessions. For each session $t\in\{1,...,T\}$, the coreset is defined as $\phi_t$ having $|\phi_t|=m^t\times N_{core}$, where $N_{core}$ represents the number of samples retained per class for the coreset. We define $\phi_{0:t}=\phi_0\cup ... \cup \phi_t$, where we set $\phi_0=\emptyset$.
%For the 

%Feature extractor is expressed as $f_{\theta}$, where $\theta$ represents network parameters. 
%The input is encoded to $\mathbb{R}^D$embedded space by passing the feature extractor. Each class $\textit{c}$ has classifier vector $w_c\in\mathbb{R}^D$ for the classification.
%We denote $w_{\mathcal{C}}=\{w_i$  for $i\in\mathcal{C}\}$ as the classifiers of each class.

\section{Method}
In this section, we introduce the overall training procedures which are illustrated in Fig. \ref{fig:detailed_schematics}. 
The following steps are proceeded for each session.
Note the commonalities and differences between base and incremental sessions for each step.
We use session index $t\in\{1,...,T\}$, where $t=1$ denotes the base session and else denotes the incremental sessions.

%\subsection{Base Session Training}
%\textbf{Initialization of class-wise token embeddings}
\subsection{Initialization of class-wise token embeddings}
We utilize learnable class-wise token embeddings which is used to generate a textual classifier for each class.
For the initialization of embeddings within the session $t$, we first consider the basic prompt $P^t_i$ = ‘a photo of a $[CLS]^t_i$', where $[CLS]^t_i$ is a class name corresponding to the class label $y^t_i$ for $i\in\{1,...,m^t\}$.
Then the class-wise token embedding of each class is initialized as 
\begin{equation}
\begin{split}
E^t_i=h(P^t_i),
\end{split}
\label{eq:clswise_base_init}
\end{equation}
where $E^t_i\in\mathbb{R}^{L(P^t_i)\times D}$.
Note that the length of encoded token for each class may differ in length, which is why we use length function $L(\cdot)$ to specify the dimension.

%\noindent\textbf{Image-object-specific prompt generation}
\subsection{Image-object-specific prompt generation}
To reflect the object-specific information within the image to the textual classifiers, 
we introduce an image-object-specific prompt generation module.
With the input image and class label $(x,y)\in \mathcal{D}^t_{tr}$, we first obtain image feature 
as
\begin{equation}
\begin{split}
f(x) = f_I(x),
\end{split}
\label{eq:img_feature}
\end{equation}
%Note that the samples from the coreset do not need to pass the image encoder since we save the image feature instead of the image itself for the coreset. The details of the coreset will be introduced more thoroughly in the subsequent sections.
Then we define the following key-map function to obtain a key incorporating information of image feature:
\begin{equation}
\begin{split}
k_x = g_\theta(f(x)),
\end{split}
\label{eq:key-map}
\end{equation}
where $k_{x}\in\mathbb{R}^D$. 
%Also, we matched the key-map output dimension with the CLIP encoder output dimension, to use initialization method shown in the initialization section.
Then we set the number of key-prompt pairs to catch the object-specific features within the session.
For each session, we construct key-prompt pairs as $\{k^t_{i}, Pr^t_{i}\}_{i=1}^{N^t}$, where $N^t$ is the number of key-prompt pairs used for session $t$.
To create image-object-specific prompts, as each image can possess multiple object-related attributes, we calculate the similarities between $k_x$ and keys and then merge prompts associated with keys exhibiting significant similarities. We calculate the similarities between the provided $k_x$ and keys, then select indices with notable similarities using the following two-dimensional $top-K$ function with $K=K_{pr}$, as elucidated in our supplementary materials:
%To form image-object-speicific prompt, each image may have multiple object-related characteristics so we calculate the similarities between the $k_x$ and keys, then combine prompts corresponding to keys with high similarities. The similarities between the given $k_x$ and the keys are calculated and we choose indexes with high similarities using two-dimensional $top-K$ function with $K=K_{pr}$, which is described in our supplementary materials, as 

\begin{equation}
\begin{split}
\{t_j,h_j\}_{j=1}^{K_{pr}} = \underset{t',i}{\arg} \:\: top\text - K\{\{sim(k_x, k^{t'}_i)\}_{i=1}^{N^{t'}}\}_{t'=1}^t,
\end{split}
\label{eq:topk_base}
\end{equation}
%for $k\in\{1,...,N^1\}$.
where $sim(\cdot,\cdot)$ is a cosine similarity function and $K_{pr}$ is number of prompts combined for bias.
%Then we calculate the similarities 
Note that we compare the $k_x$ with the keys, including the past sessions, since the class of a new session might have an object features from the past sessions.
To combine the prompts corresponding to keys with high similarity, we achieve the weights as follows:
\begin{equation}
\begin{split}
w_j = \frac{exp({sim(k_x, k^{t_j}_{h_j})})}{\sum_{i=1}^{K_{pr}}exp(sim(k_x, k^{t_i}_{h_i}))},
\end{split}
\label{eq:topk_weight}
\end{equation}
for $j\in\{1,...,K_{pr}\}$ where $exp(\cdot)$ is an exponention function. 
Note that a softmax function is employed to make the sum of weights to $1$.
Finally, the image-object-specific prompt is generated as follows:
\begin{equation}
\begin{split}
b_x = \sum_{j=1}^{K_{pr}}w_j*Pr^{t_j}_{h_j}.
\end{split}
\label{eq:image-object-specific prompt}
\end{equation}

\subsection{Initialization of key-prompt pairs}
\label{init_keypr}
%\noindent\textbf{Initialization of key-prompt pairs}
In this section, we introduce our initialization method for key-prompt pairs.
Through empirical observation, we identified that training the parameters initialized randomly only with few samples during the incremental sessions leads to challenges in updating the model to a convergence point with satisfactory performance.
Thus we search for the initialization method instead of using random values.
Given our aim to capture the inherent objective features within each class through key-prompt pairs, it is anticipated that leveraging class-wise token embeddings would likely result in successful convergence.
%Since our objective for key-prompt pairs is to catch the objective features inherent within the classes, we expect that utilizing the class-wise token embeddings may lead to successful convergence. 
So given $N^t$ as the number of key-prompt pairs used for session $t\in\{1,...,T\}$, we initialized the key-prompt pair as 
\begin{equation}
\begin{split}
k^t_i=f_T(E^t_j), Pr^t_i=E^t_j,
\end{split}
\label{eq:keyprompt_base_init}
\end{equation}
where $i\in\{1,...,N^t\}$ denotes the index for a pair and $j\in\{1,...,m^t\}$ is a randomly chosen class index from the session $t$.
As demonstrated through our experiments, our initialization method exhibits exceptional performance.
Additionally, we observed that the random selection of indices did not significantly influence the outcomes, 
as shown with further ablation studies in our supplementary materials. 
This is attributed to the crucial aspect of initializing vectors within the prompt space formed by the text encoders instead of the entire embedded space.  The primary factor impacting overall performance was the number of pairs $N^t$.
Note that we also applied the proposed key-prompt initialization method for the base session with ample training set since the results were lower when they were randomly initialized, as shown in the experimental section.

%As shown with experiments, our initialization method shows superb performances in both base and incremental sessions.

\subsection{Image-object-specific textual classifiers}
%\textbf{Image-object-specific textual classifiers}
With the class-wise token embeddings and generated image-object-specific bias
$b_x$, we now generate the textual classifiers.
Textual classifiers for each class is generated as 
\begin{equation}
\begin{split}
g^{t'}_i = f_T(b_x+E^{t'}_i),
\end{split}
\label{eq:base_textual_clf}
\end{equation}
for session index $t'\in\{1,...,t\}$ and class index $i\in\{1,...,m^{t'}\}$.

%\textbf{Training with loss function}
\subsection{Training using the loss function}
After achieving the textual classifiers, we now calculate the loss function to update the model. We first calculate logits for all currently seen classes.
For each image $(x,y)\in\mathcal{D}_{tr}^t$ and session index $t'\in\{1,...,t\}$,  
the probability logit for each class is calculated using softmax function as
\begin{equation}
\begin{split}
p(y^{t'}_i|x) = \frac{exp(sim(f(x), g^{t'}_i))}{\sum_{t'=1}^{t}\sum_{j=1}^{m^{t'}} exp(sim(f(x), g^{t'}_j))},
\end{split}
\label{eq:base_logit}
\end{equation}
for class index $i\in\{1,...,m^{t'}\}$.
Next, we employ the following cross-entropy function to compute the loss for the update:
%Then the we use cross-entropy function to calculate loss for the update as following
\begin{equation}
\begin{split}
\mathcal{L}_{CE}(x, y)= -\sum_{t'=1}^{t}\sum_{i=1}^{m^{t'}}\mathbb{1}[y^{t'}_i==y]log(p(y^{t'}_i|x)),
\end{split}
\label{eq:base_loss}
\end{equation}
where $\mathbb{1}(\cdot)$ is an indicator function.
Finally, the model is updated via the calculated loss.
We distinguish the explanations of base and incremental sessions to emphasize the difference in training.

\noindent\textbf{Base session.}
In the base session, based on the sufficient amount of training data and relatively large number of classes, the 
key-map network is trained to extract effective key from the image to compare with the key-prompt pairs, resulting to form IOS bias for the text classifier generation.
Also, the network modules corresponding to the base session are updated (e.g. key-prompt pair and class-wise token embedding). The model is updated to optimize the following equation:
\begin{equation}
\begin{split}
\underset{\theta,E^1,k^1,Pr^1}{\min} \mathds{E}_{(x,y)\sim\mathcal{D}^1_{tr}}[\mathcal{L}_{CE}(x,y)],
\end{split}
\label{eq:optimize_base}
\end{equation}
where $\theta$ is from Eq. \ref{eq:key-map}.
$E^1$, $k^1$ and $Pr^1$ denote $\{E^1_i\}_{i=1}^{m^1}$, $\{k^1_i\}_{i=1}^{m^1}$, and $\{Pr^1_i\}_{i=1}^{m^1}$, respectively.

\noindent\textbf{Incremental sessions.}
The challenge of training during incremental sessions stems from the lack of training data for past sessions, resulting in severe overfitting with high scores for the new session but poor performance on previous ones. Conversely, efforts to minimize network updates to preserve performance on past sessions have not yielded satisfactory results for new sessions \cite{yoon2023balanced, zhou2022forward, zhang2021few}.
To address these issues, we meticulously designed the update targets during incremental session training. This approach involves updating only the network modules pertinent to the current session, as detailed in the following equation:
\begin{equation}
\begin{split}
\underset{E^t,k^t,Pr^t}{\min} \mathds{E}_{(x,y)\sim\mathcal{D}^t_{tr}}[\mathcal{L}_{CE}(x,y)],
\end{split}
\label{eq:optimize_inc}
\end{equation}
where $E^t$, $k^t$ and $Pr^t$ denote $\{E^t_i\}_{i=1}^{m^t}$, $\{k^t_i\}_{i=1}^{m^t}$, and $\{Pr^t_i\}_{i=1}^{m^t}$, respectively.
Note that we exclusively update the parameters of the current session to retain knowledge from previous sessions and minimize forgetting.

\begin{table*}[!t]
\sisetup{table-format=4.0} % integer values only, up to 4 digits
    %\begin{table*}[!t]
    %% increase table row spacing, adjust to taste
    \renewcommand{\arraystretch}{1.1}    
    % if using array.sty, it might be a good idea to tweak the value of
    % \extrarowheight as needed to properly center the text within the cells
    \centering
    \begin{adjustbox}{max width=0.9\textwidth}
    \begin{tabular}{l l c c c c c c c c c c l }
    \thickhline
    \multirow{2}{*}{\textbf{Method}} & \multirow{2}{*}{\textbf{Backbone}} & \multicolumn{9}{c}{\textbf{Acc. in each session$\uparrow$ (\%) }} & \multirow{2}{*}{\textbf{AVG($\uparrow$)}}& \multirow{2}{*}{\textbf{PD($\downarrow$)}} \\ 
    \cline{3-11}
     & &  1 & 2 & 3 & 4 & 5 & 6 & 7 & 8 & 9 \\
    \hline
     %TOPIC\cite{tao2020few} & 61.31 & 50.09 & 45.17 & 41.16 & 37.48 & 35.52 & 32.19 & 29.46 & 24.42 & 36.89 \\
     %IDLVQ-C\cite{chen2020incremental} &  64.77 & 59.87 & 55.93 & 52.62 & 49.88 & 47.55 & 44.83 & 43.14 & 41.84 & 22.93 \\
    %\hline
    %CEC\cite{zhang2021few} & 72.00 & 66.83 & 62.97 & 59.43 & 56.70 & 53.73 & 51.19 & 49.24 & 47.63 & 24.37 \\
    %\hline
    %Data replay\cite{liu2022few} & 71.84 & 67.12 & 63.21 & 59.77 & 57.01 & 53.95 & 51.55 & 49.52 & 48.21 & 23.63 \\
    %MCNet\cite{ji2022memorizing} & 72.33 & 67.70 & 63.50 & 60.34 & 57.59 & 54.70 & 52.13 & 50.41 & 49.08 & 23.25 \\
    %LIMIT\cite{zhou2022few} & 72.32 & 68.47 & 64.30 & 60.78 & 57.95 & 55.07 & 52.70 & 50.72 & 49.19 & 23.13 \\
    FACT\cite{zhou2022forward} & ResNet18 & 72.6 & 69.6 & 66.4 & 62.8 & 60.6 & 57.3 & 54.3 & 52.2 & 50.5 & 60.7 & 22.1 \\
    %CLOM\cite{zou2022margin} & 73.08 & 68.09 & 64.16 & 60.41 & 57.41 & 54.29 & 51.54 & 49.37 & 48.00 & 25.08 \\
    %C-FSCIL\cite{hersche2022constrained} & 76.40 & 71.14 & 66.46 & 63.29 & 60.42 & 57.46 & 54.78 & 53.11 & 51.41 & 24.99\\
    %\hline
    ALICE \cite{peng2022few} & ResNet18 & 80.6 & 70.6 & 67.4 & 64.5 & 62.5 & 60.0 & 57.8 & 56.8 & 55.7 & 64.0 & 24.9 \\ 
    %SAVC \cite{song2023learning} & ResNet18 \\
    %SAVC \cite{song2023learning} & ResNet18 & 
    BSC \cite{yoon2023balanced} & ResNet18 & 78.4 &  74.6 & 71.4 & 68.7 & 65.9 & 63.6 & 62.0 & 60.3 & 58.5 & 67.1 & 19.8 \\
    ALICE \cite{peng2022few}* & ViT-B & 85.7 & 77.3 & 73.1 & 71.5 & 68.9 & 65.1 & 62.3 & 59.9 & 58.1 & 69.4 & 30.6 \\ 
    %SAVC \cite{song2023learning} & ViT-B \\    
    BSC \cite{yoon2023balanced}* & ViT-B & 85.2 & 80.4& 76.4& 73.2& 70.0& 67.2& 64.5& 62.7& 61.5 & 71.6 & 26.7 \\        
    %{\textbf{ILAR}} & 71.32 & 67.477 & 63.486 & 59.493 & 56.425 & 53.341 & 51.2 & 49.326 & 47.53 & \textbf{23.79}\\
    \hline
    \hline
    %CLIP ZSL* (ResNet101) & 81.8 & 81.1 & 79.6 & 79.8 & 79.0 & 78.8 & 79.3 & 79.9 & 80.1 & 79.9 & 1.7\\
    CLIP ZSL* & ViT-B/16 & 85.8 & 85.5 & 84.9 & 84.8 & 84.2 & 84.2 & 84.0 & 83.9 & 83.7 & 84.6 & 2.1 \\
    %AttriCLIP
    SV-T \cite{qiu2023semantic} &ViT-B/16 & 90.6 & 89.2 & 86.8 & 85.4 & 84.8 & 83.4 & 81.9 & 81.9 & 81.7 & 85.1 & 8.9\\
    %CPE & 90.23 & 89.56 & 87.42 & 86.80 & 86.51 & 85.08 & 83.43 & 83.38 & 82.77 \\
    CoCoOp \cite{zhou2022conditional}* &ViT-B/16 & 94.2& 91.7& 88.9& 87.8& 86.3& 84.3& 82.5& 81.9& 81.3 & 86.5 & 12.9 \\
    \hline
    %\textbf{Ours} (ResNet101) & 5 & \cmark & \cmark & MLP1  & 87.1 & 81.5 & 79.6 & 78.1 & 76.8 & 74.3 & 74.2 & 74.1 & 73.3 & 77.7 & 13.8\\     
    %\textbf{Ours} (ResNet101) & 91.2 & 87.3 & 86.4 & 86.2 & 85.4 & 85.1 & 85.0 & 84.3 & 84.0 & 86.1 & 7.2\\     
    %\textbf{Ours} (ViT-B/16) &  95.6 & 93.7 & 90.0 & 86.7 & 85.4 & 83.7 & 84.2 & 83.9 & 83.1 & 87.4 & 12.5 \\ 
    %\textbf{Ours} & 5 & \cmark & \cmark & MLP2  & \\ 
    %\textbf{Ours} (ViT-B/16)& 5 & \xmark & \cmark & MLP1  & 91.3 & 85.2 & 81.3 & 76.2 & 73.4 & 68.4 & 62.3 & 59.3 & 56.2 & 72.6 & 35.1 \\ 
    %\textbf{Ours} & 5 &  & & 95.5 & 94.1 & 91.6 & 89.8 & 88.6 & 87.5 & 87.4 & 86.9 & 86.4 \\ 
    %\textbf{Ours} (ViT-B/16)& 5 & \cmark & \xmark & MLP1  & 21.5 & 5.3 & 4.3 & 3.8 & 4.0 & 3.9 & 3.7 & 3.6 & 3.6 & 5.9 & 17.1  \\
    %\textbf{Ours} (ViT-B/16)& 5 & \cmark & \cmark & MLP2  & 95.5 & 94.1 & 91.6 & 91.2 & 89.5 & 88.9 & 89.2 & 88.6 & 88.3 & 90.8 & 7.2\\ 
    %\textbf{Ours} (ViT-B/16)& 5 & \cmark & \cmark & RES2  & 95.5 & 94.4 & 93.3 & 92.2 & 91.1 & 90.6 & 90.6 & 90.4 & 89.9 & 92.0 & 5.6\\ 
    %\textbf{Ours} (ResNet50x4) & 5 & \cmark & \cmark & MLP2  & \\     
    \textbf{Ours} &ViT-B/16 & 95.4 & 94.4 & 93.4 & 93.1 & 92.1 & 91.4 & 90.8 & 90.0 & \textbf{\underline{89.1}} & \textbf{\underline{92.2}} & 6.3\\ 
    \thickhline
    \end{tabular}
    \end{adjustbox}
    %\caption{Results of comparative studies on miniImageNet dataset with 5-way 5-shot settings}
    \label{table:sota_miniimagenet}
    %\end{table*}
%\end{subtable}
%\bigskip
%\newline
\vspace*{0.1 mm}

\caption{Comparison with state-of-the-art methods on the miniImageNet dataset in a 5-way 5-shot setting. $\uparrow$ means higher is better, while $\downarrow$ denotes lower is better. The double horizontal line distinguish whether the CLIP model was used. 
* mark denotes the re-implemented results, where we re-implemented the CLIP ZSL (zero-shot learning) and CoCoOp for the FSCIL task. Also, we re-implemented ALICE and BSC using ViT-B backbone. 
Note that we only included results that clearly specify the type of CLIP model used. Full comparison table is in our supplementary materials.
%, as larger CLIP models inherently exhibit superior performance.%; hence, consistent use of the same model type is necessary for fair comparison. 
%The 'Coreset' column indicates the number of features saved for each class, 'cls-wise' specifies whether cls-wise token embedding was used, 'Init.' denotes whether the initialization trick was applied, and 'Key-map' refers to the model type for the key map.
%Comparison with the state-of-the-art methods on miniImageNet dataset with 5-way 5-shot setting. $\uparrow$ means the higher is the better, while $\downarrow$ denotes the lower is the better. Horizontal double line distinguish whether the CLIP model was used. The result of CLIP ZSL (zero-shot learning) and CoCoOp is re-implemented for the FSCIL task. Note that we only included the results that clearly clarified the type of CLIP model used, since the bigger CLIP model necessarily shows better performance so consistent use of the model type is needed for the fair comparison. Column `Coreset denotes the number of feature saved for each class, `cls-wise' denotes whether cls-wise token embedding was used, `Init.' denotes whether the initialization trick was used, and `Key-map' denotes the model type for key map.
%Unreported results with no codes available are excluded from the table. 
}
\label{table:sota}
\end{table*}

\begin{figure*}[t]
    \centering 
    \includegraphics[width=0.9\linewidth]{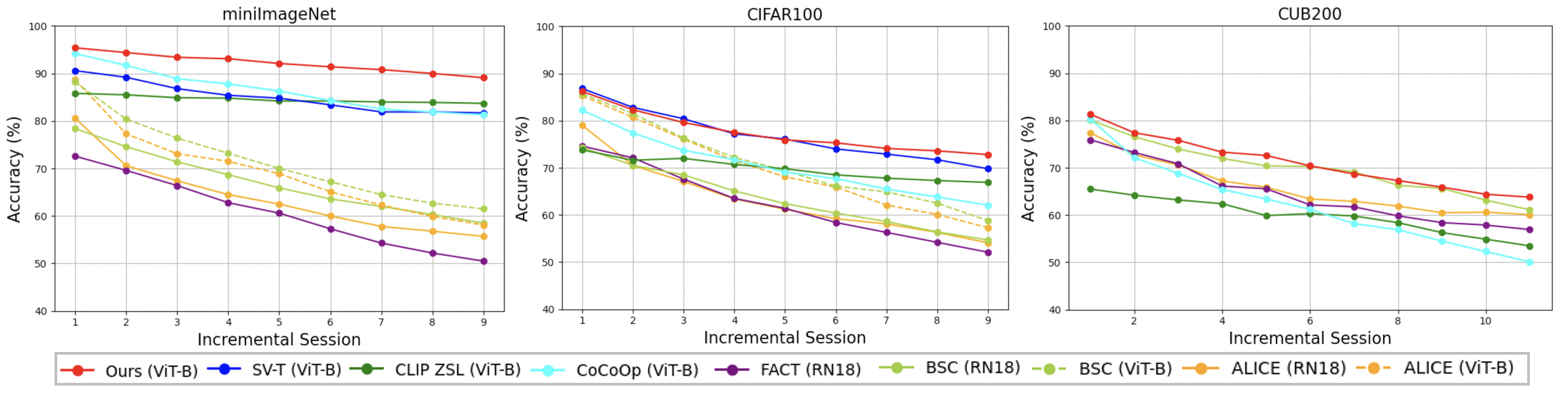}
    \caption{
    Comparison with the state-of-the-art methods on diverse datasets.
    Note that for the CUB200 dataset, the previous methods without using the CLIP model utilize ImageNet-pretrained model due to the difficulty of fine-grained dataset.
    }
    \label{fig:sota}
\vspace{-2mm}%
\end{figure*}

\section{Experiments}
\label{sec:exps}

\subsection{Experimental Setup}
\label{sec:exp_setups}

\textbf{Dataset.}
We conducted performance evaluations of the comparative methods across three widely-used datasets: miniImageNet \cite{vinyals2016matching}, CUB200 \cite{wah2011caltech}, and CIFAR100 \cite{krizhevsky2019cifar}. The miniImageNet dataset, a 100-class subset of ImageNet-1k \cite{russakovsky2015imagenet}, is commonly employed in few-shot learning. 
%It comprises 100 classes, each featuring 
It contains 500 training images and 100 test images for each class, with each image measuring $84\times84$ pixels. CIFAR100 offers 100 classes, each containing 500 training images and 100 test images, with each image having dimensions of $32\times32$ pixels. The CUB200 dataset %, designed for fine-grained image classification, was introduced for incremental learning and 
includes around 6,000 training images and 6,000 test images spread over 200 bird classes, having $224\times224$ pixel size.
%These images are resized to $256\times256$ pixels and then cropped to $224\times224$ pixels for training purposes.

In line with the experimental design of previous work, we randomly chose 60 classes as base classes for miniImageNet and CIFAR100, dividing the remaining incremental classes into eight sessions, each encompassing five classes. For CUB200, we split the data into 100 base classes and 100 incremental classes, with the incremental classes further divided into ten sessions of ten classes each. For all datasets, we used five training images for each class in our 5-shot experiments, and we utilized all test images to assess generalization performance and avoid overfitting.

\noindent\textbf{Evaluation protocol.}
% 그러나 FSCIL에서 흔히 사용된 metric은 모든 세션 성능 평가에 base session이 들어갔고, 심지어 그 class 수의 비중이 new에 비해 훨씬 크다는 특징이 있었기에 base session 성능을 잘 지키는 것만으로도 좋은 지표를 얻을 수 있었음.
% 또한 empirically new task 학습 시 FE 변경을 위한 유의미한 lr을 주면 base session acc가 현저히 떨어진다는 한계가 있었음.
% 따라서 많은 연구들은 FE를 inc session에서 freeze하거나 낮은 learning rate를 주었고, 때문에 new task만으로 구성된 평가에선 매우 낮은 성적을 보였음.
%However, current metric for FSCIL measures the classification accuracy for the total classes seen so far at each session. 
%This obviously includes base session classes, which occupy relatively large portion compared to new classes.
%Thus, just maintaining accuracy for the base session classes resulted high scores.
%Actually many researches freeze or give infinitesimal learning rates to the feature extractors in the incremental sessions.
%Eventually updates for learning knowledge for the new classes were limited, resulting poor accuracy scores for the new classes.
Prior research primarily used the average accuracy (AVG) and performance dropping (PD) as the key metric to demonstrate the efficacy of the algorithms. AVG is the average performance over all sessions and PD is defined as the absolute difference in test accuracy between the base session and the final session. To elaborate, we denote $\mathcal{A}^t_\mathcal{Y}$ as the accuracy assessed solely on the test sets of classes 
$\mathcal{Y}$. Here, probabilities are calculated among classes $\mathcal{Y}^{1:t}$, using the network model trained after the $t^{th}$ session.
Then, AVG and PD are respectively defined as
\begin{equation}
\begin{split}
AVG = \frac{\sum_{t=1}^T\mathcal{A}^t_{\mathcal{Y}^{1:t}}}{T}, \:\:
PD = \mathcal{A}^1_{\mathcal{Y}^1}-\mathcal{A}^T_{\mathcal{Y}^{1:T}}.
\end{split}
\label{eq:PDAVG}
\end{equation}

To precisely measure the performance of the model during the sessions on classes of base and incremental sessions, we further use the following metrics \cite{yoon2023balanced} to measure the new task learning ability (NLA) and base task maintaining ability (BMA):
\begin{equation}
\begin{split}
NLA = \frac{\sum_{t=2}^T\mathcal{A}^t_{\mathcal{Y}^{2:t}}}{T-1}, \;\;
BMA = \frac{\sum_{t=1}^T\mathcal{A}^t_{\mathcal{Y}^{1}}}{T}.
\end{split}
\label{eq:NLABMA}
\end{equation}

\noindent\textbf{Implementation details.}
In the experiments, we utilized the ViT-B/16 model \cite{dosovitskiy2020image} as the backbone for CLIP.
We used 5 for base and 3 for incremental sessions as training epochs. 
%$N_{core}$ is aligned with the number of shots in each incremental session. 
Default top-K value used for key-prompt pair was 3, and the number used for key-prompt pairs were 20 and 3 for base and incremental sessions, respectively.
Note that all values reported in this paper are average of results from 10 individual runs to reduce the randomness.
Further details are described in our supplementary materials.

\subsection{Main Results}

\textbf{Comparison with the state-of-the-art methods.}
We compared our proposed method with the state-of-the-art FSCIL methods. 
Considering that the utilization of the CLIP model inherently results in improved performance, we distinguished between methods employing the CLIP model in Table \ref{table:sota}.
%Furthermore, due to the absence of the commonly used ResNet18 backbone in the CLIP model, we chose to use the ViT-B backbone for CLIP. This led us to re-implement previous methods using the same backbone.
 Additionally, since the CLIP model does not offer the commonly-used ResNet18 backbone found in previous methods, we chose to use the ViT-B backbone for CLIP.
 This led us to re-implement previous methods using the same backbone.
% and subsequently re-implemented previous methods using the same backbone. 
  As evidenced in Table \ref{table:sota} and Fig. \ref{fig:sota}, our method surpasses the state-of-the-art models on the CIFAR100, miniImageNet, and CUB200 datasets.

\begin{figure}[!t]
%\begin{figure}[H] 
\centering
\begin{subfigure}[b]{7cm}
\includegraphics[width=7cm]{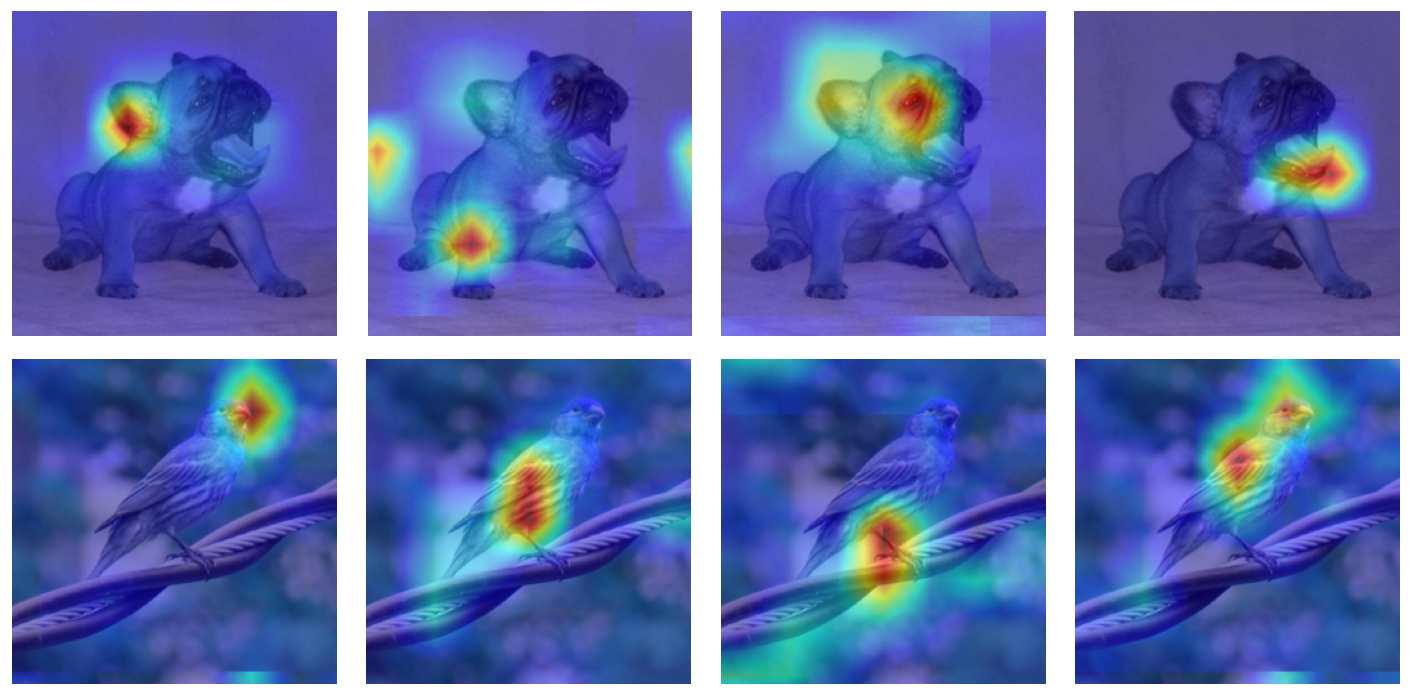}
\caption{Role of different prompts of the same image}\label{fig:gradcam_a}
\end{subfigure}
\begin{subfigure}[b]{7cm}
\includegraphics[width=7cm]{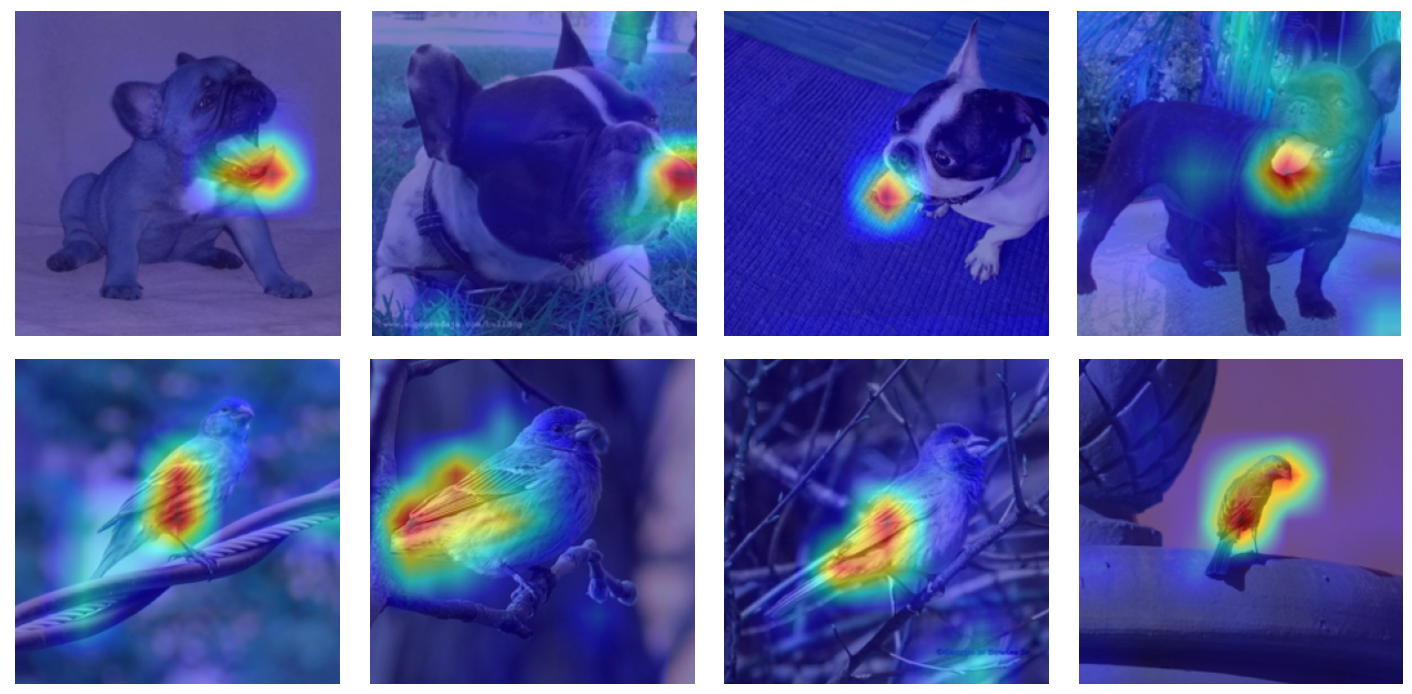}
\caption{Role of a single prompt on different images}\label{fig:gradcam_b}
\end{subfigure}
\vspace{-2mm}
\caption{Visualization of the role of the learned prompts.}
\label{fig:tsne}
\vspace{-2mm}
\end{figure}

\noindent\textbf{Visualization of the effacity of image-object-specific modules.}
%To demonstrate our claim that our key-prompts are learned to catch the image-object-specific features, we visualize the attention map using Grad-CAM \cite{selvaraju2017grad}. To emphasize the characterstics of each learned key-prompt pair, we visualize the image contents corresponding to each prompts. To highlight the contents within the image which contributed to classify for each class,  we calculate the loss between the image feature and the generated textual classifiers.
%$f_I(x)$ and the calculated textual classifiers $g^{t'}_i$, following the notations from Eq. \ref{eq:img_feature}, \ref{eq:base_textual_clf}. In order to analyze each prompt, instead of combining the prompts with high similarity as in Eq. \ref{eq:image-object-specific prompt}, we generated the classifier only using single prompt. Thus for the given image, we generated classifiers using different single prompts and achieved the attention map to analyze the prompts.
To validate our claim that our key-prompts are designed to capture image-object-specific features, we employ the attention map visualization method for transformer models \cite{chefer2021generic}. To emphasize the characteristics of each learned key-prompt pair, we display the image content that correlates with each prompt. To highlight the image content influencing the classification for each class, we compute the loss between the image feature and the generated textual classifiers. 
With the trained key-prompts using Eq. \ref{eq:image-object-specific prompt}, 
we analyze the role of each prompt. 
In Fig. \ref{fig:gradcam_a}, different columns correspond to distinct prompts from the selected $top-K$ prompts extracted from the given image.
 We could see that different prompts encapsulate diverse object attributes. To verify the functionality of each prompt, we apply a single prompt across diverse images, as shown in Fig. \ref{fig:gradcam_b}.
We observe that attention from the same prompt highlights similar object features. Moreover, we observed that the majority of these features are related to the objects themselves rather than the background. 
%discerned that most of these features pertain to the objects rather than the background.
%Only objects related. Incremental session feature

\begin{figure}
\centering
\includegraphics[width=0.85\linewidth]{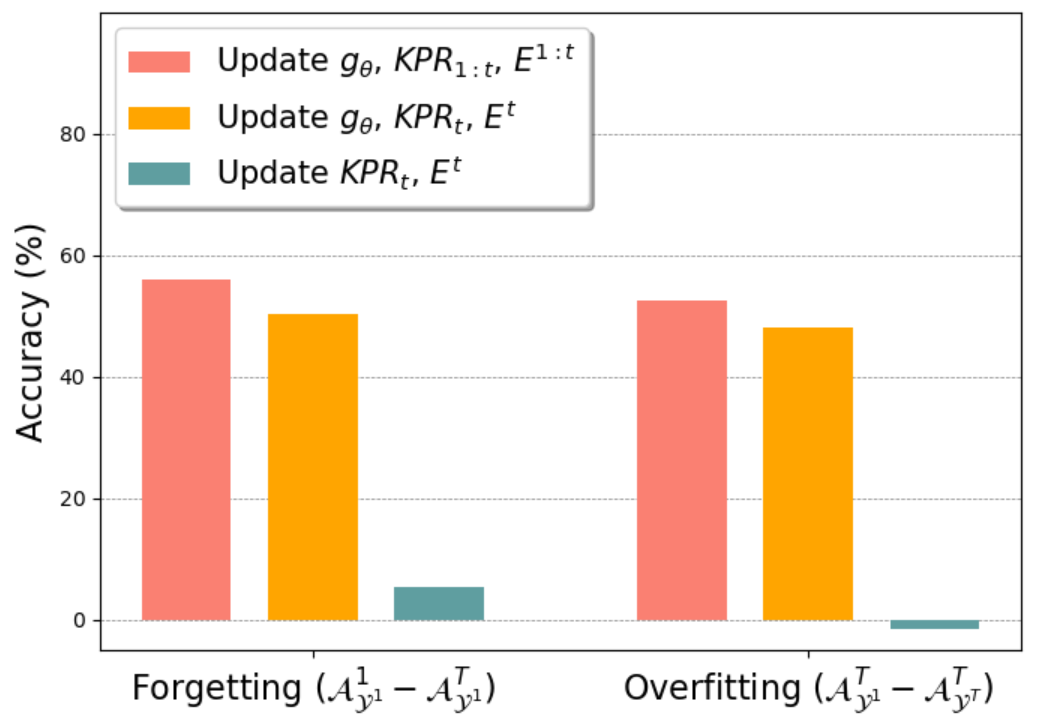}
\vspace{-2mm}
\caption{Effects of the parameter update range during incremental session training on forgetting and overfitting. The parameter notation follows from Fig. \ref{fig:detailed_schematics}. To assess forgetting, we compare the base session performance of the model after the base session training with that after the final session. To evaluate overfitting, we compare the performance on both the base and final sessions after the final session training.}
\label{fig:new_train_ablation}
\vspace{-2mm}
\end{figure}

\begin{table}[!t]
\renewcommand{\arraystretch}{1.1}
% if using array.sty, it might be a good idea to tweak the value of
% \extrarowheight as needed to properly center the text within the cells
\setlength\tabcolsep{4pt}
\centering
\resizebox{0.9\columnwidth}{!}{%
\begin{tabular}{c | c c | c c}
\thickhline
Loss &  AVG$(\uparrow)$ & PD$(\downarrow)$ & NLA$(\uparrow)$ & BMA$(\uparrow)$\\
\hline
\hline
CEC* \cite{zhang2021few} & 57.7 & 24.6 & 18.6 & 68.3\\
\hline
FACT* \cite{zhou2022forward} & 60.7 & 22.1 & 13.5 & 75.2 \\
\hline
BSC* \cite{yoon2023balanced} & 68.7 & 20.2 & 37.8 & 77.8 \\
\hline
CLIP ZSL* & 84.6 & 2.1 & 85.5 & 84.2\\
\hline
CoCoOp* & 86.5 & 12.9 & 86.1 &  86.9  \\
\hline
Ours & 92.2 & 6.3 & 91.1 & 92.4 \\
\thickhline
\end{tabular}
}
\caption{Comparison of methods on miniImageNet with additional metrics. * denotes the re-implemented results.
%In order to monitor the validity of the extracted represen-
%tations in both the base session and the incremental session,
%which are the purpose of the generic representation, NLA
%and BMA metrics were additionally used.
}
\label{table:furthermetrics}
\vspace{-3mm}
\end{table}

\begin{table*}[!t]
\sisetup{table-format=4.0} % integer values only, up to 4 digits
    \renewcommand{\arraystretch}{1.1}
    \centering
    \begin{adjustbox}{max width=\textwidth}
    \begin{tabular}{l l l l l c c c c c c c c c c c c l }
    \thickhline
    \multirow{2}{*}{\textbf{Method}}  & \multirow{2}{*}{Init.}  & \multirow{2}{*}{\shortstack{cls-\\wise}} & \multirow{2}{*}{\shortstack{Key\\map}} & \multicolumn{9}{c}{\textbf{Acc. in each session$\uparrow$ (\%) }} & \multirow{2}{*}{\textbf{AVG($\uparrow$)}}& \multirow{2}{*}{\textbf{PD($\downarrow$)}} & \multirow{2}{*}{\textbf{NLA($\uparrow$)}} & \multirow{2}{*}{\textbf{BMA($\uparrow$)}} \\ 
    \cline{5-13}
     & & & & 1 & 2 & 3 & 4 & 5 & 6 & 7 & 8 & 9 \\
     \hline
    %\textbf{Ours} (ViT-B/16) & \cmark & \cmark  & FC2  & 95.6 & 93.7 & 90.0 & 86.7 & 85.4 & 83.7 & 84.2 & 83.9 & 83.1 & 87.4 & 12.5 \\ 
    \textbf{Ours} (ViT-B/16)& \cmark & \xmark & FC1  & 91.3 & 85.2 & 81.3 & 76.2 & 73.4 & 68.4 & 62.3 & 59.3 & 56.2 & 72.6 & 35.1 & 57.3 & 76.2\\ 
    \textbf{Ours} (ViT-B/16)& \xmark & \cmark & FC1  & 91.5 & 76.5 & 65.2 & 59.2 & 51.2 & 48.7 & 44.2 & 41.8 & 39.5 & 57.5 & 52.0 & 15.7 & 71.3 \\
    \textbf{Ours} (ViT-B/16)& Inc. & \cmark & FC1  &  91.5 & 90.2 & 89.5 & 89.1 & 88.3 & 87.7 & 86.9 & 85.8 & 84.9 & 88.2 & 6.6 & 86.3 & 89.8 \\
    \textbf{Ours} (ViT-B/16)& \cmark & \cmark & FC2  & 95.5 & 94.1 & 92.6 & 91.2 & 89.5 & 89.2 & 88.9 & 88.6 & 88.3 & 90.9 & 7.2 & 89.8 & 91.2\\ 
    \textbf{Ours} (ViT-B/16)& \cmark & \cmark & RES2  & 95.5 & 94.2 & 93.1 & 92.1 & 91.1 & 90.3 & 89.9 & 88.8 & 88.6 & 91.5 & 6.9 & 90.3 & 91.8\\ 
    %CLIP ZSL* (ResNet101) & 81.8 & 81.1 & 79.6 & 79.8 & 79.0 & 78.8 & 79.3 & 79.9 & 80.1 & 79.9 & 1.7\\
    \textbf{Ours} (ResNet101) & \cmark & \cmark & FC1 & 91.2 & 87.3 & 86.4 & 86.2 & 85.4 & 85.1 & 85.0 & 84.3 & 84.0 & 86.1 & 7.2 & 83.5 & 88.5\\    
    \textbf{Ours} (ViT-B/16) & \cmark & \cmark & FC1  &95.4 & 94.4 & 93.4 & 93.1 & 92.1 & 91.4 & 90.8 & 90.0 & \textbf{\underline{89.1}} & \textbf{\underline{92.2}} & \textbf{\underline{6.3}} & \textbf{\underline{91.1}} & \textbf{\underline{92.4}}\\ 
    \thickhline
    \end{tabular}
    \end{adjustbox}
    \label{table:sota_miniimagenet}
\vspace*{0.1 mm}
\caption{Ablation study of the proposed method on miniImageNet dataset in a 5-way 5-shot setting. `cls-wise' specifies whether cls-wise token embedding was used, `Init.' denotes whether the initialization trick was applied (`Inc.' means that the trick is applied only for incremental sessions), and `Key-map' refers to the model type for the key map.
FC1 and FC2 denotes the fully connected layer with one and two layers, while RES2 denotes FC2 with residual path added. 
The detailed structures are described in our supplementary materials.
%Comparison with the state-of-the-art methods on miniImageNet dataset with 5-way 5-shot setting. $\uparrow$ means the higher is the better, while $\downarrow$ denotes the lower is the better. Horizontal double line distinguish whether the CLIP model was used. The result of CLIP ZSL (zero-shot learning) and CoCoOp is re-implemented for the FSCIL task. Note that we only included the results that clearly clarified the type of CLIP model used, since the bigger CLIP model necessarily shows better performance so consistent use of the model type is needed for the fair comparison. Column `Coreset denotes the number of feature saved for each class, `cls-wise' denotes whether cls-wise token embedding was used, `Init.' denotes whether the initialization trick was used, and `Key-map' denotes the model type for key map.
%Unreported results with no codes available are excluded from the table. 
}
\label{table:ablation_modules}
\end{table*}

\begin{table}[h]
\centering
\resizebox{0.9\columnwidth}{!}{%
\begin{tabular}{c c | c c c c}
\thickhline
$N^1$ & $N^t (t>1)$ & AVG$(\uparrow)$ & PD$(\downarrow)$ & NLA$(\uparrow)$ & BMA$(\uparrow)$ \\ \hline
20  & 1 & 91.9 & 6.7 & 90.2 & 92.1 \\ \hline
20  & 3 & 92.2 & 6.3 & 91.1 & 92.4\\ \hline
40  & 3 &  92.1 &  6.4 & 90.8 & 92.2 \\ \hline
60  & 5 &  90.0 & 10.5 & 88.5 & 90.4 \\ \hline
100  & 10 & 86.3 & 14.2 & 84.2 & 87.2 \\ \hline
\thickhline
\end{tabular}
}
\caption{Ablation study on the number of key-prompt pairs $N^t$ for base and incremental sessions.}
\label{tab:ablation_keyprompt}
\vspace{-3mm}
\end{table}

\begin{table}[h]
\centering
\resizebox{0.8\columnwidth}{!}{%
    \begin{tabular}{c | c c c c}
    \thickhline
    top-K  & AVG$(\uparrow)$ & PD$(\downarrow)$ & NLA$(\uparrow)$ & BMA$(\uparrow)$  \\ \hline
    1 & 91.8 & 6.7 & 90.3 & 92.0 \\ \hline
    2 & 92.0 & 6.5 & 90.8 & 92.3  \\ \hline
    3 & 92.2 & 6.3 & 91.1 & 92.4 \\ \hline
    4 & 92.1 & 6.4 & 90.7 & 92.4  \\ \hline
    5 & 91.3 & 7.5 & 90.0 & 91.5 \\ \thickhline
    \end{tabular}
    }
\caption{Ablation study on top-K value used for prompt bias generation.}
\label{tab:ablation_topk}
\vspace{-3mm}
\end{table}

\noindent\textbf{Effectiveness on reducing the overfitting and the forgetting.}
During incremental session training, our learning scheme was tailored to minimize forgetting past session knowledge and overfitting to the new session. To examine the impact of our update scheme on overfitting and forgetting, we conducted an ablation study, as depicted in Fig. \ref{fig:new_train_ablation}. The findings highlight the significance of the choice of parameters to be updated. Specifically, updating the key-map network, which is utilized for both base and incremental classes, solely with current session training data led to substantial forgetting and overfitting. In contrast, updating only the parameters related to the current session yielded the most favorable results.

\comment{
\begin{figure}
\centering
\begin{subfigure}[b]{.45\linewidth}
\includegraphics[width=\linewidth]{images/tsne_a.png}
\caption{Before the training}\label{fig:tsne_ce}
\end{subfigure}
\begin{subfigure}[b]{.45\linewidth}
\includegraphics[width=\linewidth]{images/tsne_b.png}
\caption{After the training}\label{fig:tsne_scl}
\end{subfigure}
\caption{t-SNE analysis of representations for the base and incremental session classes. (a) shows the state before incremental session training, while (b) displays the state afterward.}
\label{fig:tsne}
\end{figure}
}

Furthermore, we compared our results on the NLA and BMA metrics, respectively representing the overall ability to learn new tasks and maintain knowledge of previous sessions without forgetting, with prior FSCIL works in Table \ref{table:furthermetrics}. Our method demonstrates superb performance across all metrics, effectively reducing both forgetting and overfitting.

\subsection{Ablation Studies}
\textbf{Effects of initialization methods.}
In the preceding sections, we utilized initialization tricks for key-prompt pairs with two objectives.
Firstly, since the primary objective of key-prompt pairs is to capture class-specific object features, we anticipated that initializing with class-wise token embeddings would facilitate convergence.
Secondly, considering the limited amount of training data available in incremental sessions, initializing parameters with meaningful vector values instead of random ones could help alleviate potential issues of severe overfitting.
%and reduce the likelihood of the model converging to undesirable low-performance states. 
As seen in Table \ref{table:ablation_modules} (row 2 \textit{vs}. 7), using random initialization led to convergence failures in the incremental sessions. Our proposed initialization method significantly increased NLA (2 \textit{vs}. 3), confirming its effectiveness for learning in incremental sessions. Additionally, (3 \textit{vs}. 7) implies that applying random initialization to the base session results in a decrease in overall performance.
%As observed in Table \ref{table:ablation_modules} (row 2 \textit{vs}. 7), random initialization resulted in convergence failures for the incremental sessions. Proposed initialization method increases NLA dramatically (row 2 \textit{vs}. 3), verifying it's effectiveness for the incremental session learning. Also (3 \textit{vs}. 7) implies that applying random initialization to the base session decreases the overall performance.

\noindent\textbf{Using the classwise learnable token embeddings.}
In this paper, we used learnable token embeddings for each class, instead of using common learnable embedding as in \cite{zhou2022learning}. 
%The purpose of using class-wise embeddings is to enhance the capacity of the model to catch the features of each class.
%Using common embedding instead of class-wise results catastrophic forgetting % when the common embedding is updated 
 %during the incremental sessions, %on the other hand, results poor performance for incremental classes when is not updated, 
%as shown in the Table \ref{table:ablation_modules}.%, we can see that adopting classwise embeddings enhanced the performance by xxx, 
%The purpose of using class-wise embeddings is to enhance the capacity of the model to catch the features of each class. Using common embedding instead of class-wise results catastrophic forgetting during the incremental sessions, as shown in the Table \ref{table:ablation_modules}.
The purpose of adopting class-wise embeddings is to enhance the model's capacity to capture the unique features of each class. Using common embedding instead of class-wise results in catastrophic forgetting during the incremental sessions, as shown in Table \ref{table:ablation_modules} (1 \textit{vs}. 7).

%\noindent\textbf{Necessity of coresets.}
%Since the training set in the incremental sessions only contain data for the current session, training without the coreset for the classes of the past sessions easily causes severe overfitting. 
%So we built the coreset to prevent the overfitting in the incremental sessions.
%The results of Table \ref{table:ablation_modules} (1 \textit{vs}. 6) also demonstrate the effects of the coreset, as the classification performance steeply decrease as the sessions proceed without using the coreset.

\noindent\textbf{Network for key-map.} 
Ablation study on the key-map network structure in Table \ref{table:ablation_modules} (4,5 \textit{vs}. 7) shows that using FC2 and RES2 slightly reduced performances compared to FC1. We infer from the results that a single fully connected layer is sufficient for the key-map network structure, and larger models diminish performance.

%\noindent\textbf{Comparison on further metrics.} In Table \ref{table:furthermetrics}, we proceed to compare FSCIL methods based on additional metrics. The results show that our method shows outstanding performance on all metrics without forgetting or overfitting.

\noindent\textbf{Number of prompt-key pairs for each session.}
%We analyzed the effects of number of prompts. Our interest was that how many prompts are optimal to catch the features of the class images within the session. As seen in the Table  When the number is small Also we can see that if the number of prompts gets too big, 
%We proceed ablation study to analyze the effects of number of key-prompt pairs used to the performance.
As shown in Table \ref{tab:ablation_keyprompt}, changes in the number of key-prompt pairs resulted in slight changes to the performance. Note that using a number smaller than the total class number led to better score, which may be due to the inclusion of shared features among similar classes enhancing the performance.

\noindent\textbf{Number of the top-K values.} We further proceeded to analyze the effect of the number of $top-K$ value used in Eq. \ref{eq:topk_base} to the overall performance. 
As we can see from Table \ref{tab:ablation_topk}, using less or too much matched prompts to combine the prompt bias decreased the performance.

\section{Conclusion}
\label{sec:conclusion}
In this paper, we proposed a training architecture for the FSCIL task using image-object-specific (IOS) textual classifiers, generated through our proposed IOS prompt bias generation module. Concerning the generation module, we crafted the architecture to reduce past knowledge forgetting and current session overfitting during incremental session training. Our method demonstrated marked enhancements compared to state-of-the-art approaches, and we confirmed the presence of IOS features in the generated classifiers. Furthermore, we offer ablation studies to analyze the effects of each module used within the training architecture.

%In this paper, we proposed training architecture for FSCIL task based on utililzing image-object-specific (IOS) textual classifiers, generated with proposed IOS prompt bias generation module. With respective to the generation module, we designed the architecture to minimize the forgetting of the past knowledge and overfitting to the current session during the training in incremental sessions. Our method showed significant improvements compared to the state-of-the-art methods, and we verified that the generated classifiers showed IOS features. Also we provide ablation studies to analyze the effects of each modules used within the training architecture.
{
    \small
    \bibliographystyle{ieeenat_fullname}
    \bibliography{main}
}

\appendix
\newpage
\clearpage
\part*{Supplementary Material} 
\label{app:sec:supp}

\section{Further Details}
\subsection {Implementation Details}
We utilized 4 NVIDIA GeForce RTX 3090 GPUs for training and testing.
The model was trained by the stochastic gradient descent (SGD) optimizer \cite{ruder2016overview} with a learning rate 0.002, momentum 0.9, and decay 0.0005. The batch size was 16 for both base and incremental sessions. The overall implementation was based on Pytorch.

\subsection{Experiment Detail for Fig. \ref{fig:intro_compare_figure}}
We proceeded the FSCIL experiment on the miniImageNet dataset, with ViT-B/16 backbone for the CLIP image encoder.
The results are re-implemented values.

\subsection{Detailed Structures of the Key-map Network}
We mentioned ‘FC1', ‘FC2', and the ‘RES2' options for the key-map network in the Table \ref{table:ablation_modules}.
‘FC1' is one fully connected layer with input and output dimension of 512. 
‘FC2' is two fully connected layer with ReLU activation between the layers. 
‘RES2' is two-layer residual layer with structure as linear-ReLU-linear having residual path between the input and the output of second linear layer.

\subsection {Implementation Details}
We utilized 4 NVIDIA GeForce RTX 3090 GPUs for training and testing.
The model was trained by the stochastic gradient descent (SGD) optimizer \cite{ruder2016overview} with a learning rate 0.002, momentum 0.9, and decay 0.0005. The batch size was 16 for both base and incremental sessions. The overall implementation was based on Pytorch.

\section{Effects of Randomness During the Key-prompt Initialization}
In Section \ref{init_keypr}, we randomly selected indices within the classes of the current session for key-prompt initialization. We utilized five different random seeds and conducted 10 iterations of random key-prompt initialization for each seed. This process was carried out on the miniImageNet dataset, resulting in an average standard deviation of 0.84 for the AVG metric with the mean of 92.2. This relatively low standard deviation helps to support the reliability of our initialization method.

%In Fig. xxx, 

\section{Two-dimensional top-K function}
Given a two-dimensional function $h(x,y)$ having the integer domains $x\in\{1,...,N\}$ and $y\in\{1,...,M\}$,
we define the two-dimensional $top-K$ function as follows.
Firstly, for integer $z$, we define quotient and remainder functions as follows:
\begin{equation}
\begin{split}
Q(z) = \lfloor (z-1)/M \rfloor +1 , R(z) = z-(Q(z)-1)M
\end{split}
\label{app:eq:QR}
\end{equation}
where  $\lfloor \cdot \rfloor$ is a floor function. 
Then, we define modified a function with single attribute as  
\begin{equation}
\begin{split}
h^*(z)= h(Q(z), R(z)),
\end{split}
\end{equation}
for $z\in\{1,...,NM\}$.
With the function with a single attribute, we obtain $top-K$ index as
\begin{equation}
\begin{split}
\{z_l\}_{l=1}^{K} = \underset{z}{\arg} \:\: top\text - K\{h^*(z)\}_{z=1}^{NM}.
\end{split}
\end{equation}
Finally, the two-dimensional $top-K$ function is defined as
\begin{equation}
\begin{split}
\{Q(z_l), R(z_l)\}_{l=1}^K \equiv \underset{x,y}{\arg} \:\: top\text - K\{\{h(x,y)\}_{y=1}^{M}\}_{x=1}^N.
\end{split}
\end{equation}

\section{Pseudo-code of proposed algorithm}
Pseudo-code of the our overall algorithm is introduced in Algorithm \ref{alg:algorithm}.

\begin{algorithm}
\caption{Pseudo-code of overall learning algorithm}\label{alg:cap}
\begin{algorithmic}[1]
\Require Input image dataset $\{\mathcal{D}^t_{tr},\mathcal{D}^t_{te}\}_{t=1}^T$ , CLIP image and text encoder $f_I(\cdot)$ and $f_T(\cdot)$, random initialized key-map $g_\theta(\cdot)$.
\Ensure Optimized key-map $g_\theta(\cdot)$ and class-wise token embeddings $\{\{E^t_j\}_{j=1}^{m^t}\}_{t=1}^T$ and key-prompt pairs $\{\{k^t_i,Pr^t_i\}_{i=1}^{N^t}\}_{t=1}^T$.
\For{session $t\in\{1,...,T\}$}
%\While{$N \neq 0$}
\State Initialize class-wise embeddings $\{E^t_j\}_{j=1}^{m^t}\}$ as Eq. \ref{eq:clswise_base_init}
\State Initialize key-prompt pairs $\{k^t_i,Pr^t_i\}_{i=1}^{N^t}$ as Eq. \ref{eq:keyprompt_base_init}
\State Get train data $(x,y)\in\mathcal{D}^t_{tr}$.
\State Get image feature $f(x)$ as Eq. \ref{eq:img_feature}.
\State Get key $k_x$ as Eq. \ref{eq:key-map}.
\State Match keys and get index from top-K as Eq. \ref{eq:topk_base}.
\State Get prompt bias $b_x$ combining prompts as Eq. \ref{eq:image-object-specific prompt}.
\State Generate classifiers $\{\{g^{t'}_i\}_{i=1}^{m^{t'}}\}_{t'=1}^t$ as Eq. \ref{eq:base_textual_clf}.
\State Calculate logits $\{\{p(y^{t'}_i|x)\}_{i=1}^{m^{t'}}\}_{t'=1}^t$ as Eq. \ref{eq:base_logit}.
\State Calculate loss function as Eq. \ref{eq:base_loss}.
\State Update parameters following Eq. \ref{eq:optimize_base}.
\State Evaluate the model with $\mathcal{D}^t_{te}$.
%\EndWhile
\EndFor
\end{algorithmic}
\label{alg:algorithm}
\end{algorithm}

\section{Full Comparison Table}
In extension to the Table \ref{table:sota}, we provide full comparison table including more previous FSCIL methods and the results for PD metric.
Results on miniImageNet, CIFAR100, CUB200 datasets are each in Table \ref{app:table:sota_mini}, \ref{app:table:sota_cifar}, and \ref{app:table:sota_cub}.

\begin{table*}[!t]
\centering
\sisetup{table-format=4.0} % integer values only, up to 4 digits
    \renewcommand{\arraystretch}{1.1}    
    \centering
    \begin{adjustbox}{max width=1.0\textwidth}
    %\begin{adjustbox}{angle=90}
    \begin{tabular}{l l c c c c c c c c c c l }
    \thickhline
    \multirow{2}{*}{\textbf{Method}} & \multirow{2}{*}{\textbf{Backbone}} & \multicolumn{9}{c}{\textbf{Acc. in each session$\uparrow$ (\%) }} & \multirow{2}{*}{\textbf{AVG($\uparrow$)}}& \multirow{2}{*}{\textbf{PD($\downarrow$)}} \\ 
    \cline{3-11}
     & &  1 & 2 & 3 & 4 & 5 & 6 & 7 & 8 & 9 \\
    \hline
     TOPIC \cite{tao2020few} & ResNet18& 61.3 & 50.1 & 45.2 & 41.2 & 37.5 & 35.5 & 32.2 & 29.5 & 24.4 & 39.6 & 36.9 \\
     IDLVQ-C \cite{chen2020incremental} & ResNet18&  64.8 & 59.9 & 55.9 & 52.6 & 49.9 & 47.6 & 44.8 & 43.1 & 41.8 & 51.2 & 22.9 \\
    %\hline
    CEC \cite{zhang2021few} & ResNet18& 72.0 & 66.8 & 63.0 & 59.4 & 56.7 & 53.7 & 51.2 & 49.2 & 47.6 & 57.7& 24.4 \\
    %\hline
    Data replay \cite{liu2022few} & ResNet18& 71.8 & 67.1 & 63.2 & 59.8 & 57.0 & 54.0 & 51.6 & 49.5 & 48.2 & 58.0 &23.6 \\
    MCNet \cite{ji2022memorizing} & ResNet18& 72.3 & 67.7 & 63.5 & 60.3 & 57.6 & 54.7 & 52.1 & 50.4 & 49.1 & 58.6 & 23.3 \\
    LIMIT \cite{zhou2022few} & ResNet18& 72.3 & 68.5 & 64.3 & 60.8 & 58.0 & 55.1 & 52.7 & 50.7 & 49.2 & 59.1& 23.1 \\
    FACT \cite{zhou2022forward} & ResNet18 & 72.6 & 69.6 & 66.4 & 62.8 & 60.6 & 57.3 & 54.3 & 52.2 & 50.5 & 60.7 & 22.1 \\
    CLOM \cite{zou2022margin} & ResNet18& 73.1 & 68.1 & 64.2 & 60.4 & 57.4 & 54.3 & 51.5 & 49.4 & 48.0 & 58.5 & 25.1 \\
    C-FSCIL \cite{hersche2022constrained} & ResNet18& 76.4 & 71.1 & 66.5 & 63.3 & 60.4 & 57.5 & 54.8 & 53.1 & 51.4 & 61.6 &25.0\\
    %\hline
    ALICE \cite{peng2022few} & ResNet18 & 80.6 & 70.6 & 67.4 & 64.5 & 62.5 & 60.0 & 57.8 & 56.8 & 55.7 & 64.0 & 24.9 \\ 
    SAVC \cite{song2023learning}\dag & ResNet18 & 80.5 & 75.0 & 72.0 & 68.0 & 63.0 & 62.0 & 60.0 & 59.0 & 58.0 & 66.4 & 22.5\\    
    BSC \cite{yoon2023balanced} & ResNet18 & 78.4 &  74.6 & 71.4 & 68.7 & 65.9 & 63.6 & 62.0 & 60.3 & 58.5 & 67.1 & 19.8 \\
    ALICE \cite{peng2022few}* & ViT-B & 88.7 & 77.3& 73.1& 71.5& 68.9& 65.1& 62.3& 59.9& 58.1 & 69.4 & 30.6 \\
    %SAVC \\    
    BSC \cite{yoon2023balanced}* & ViT-B &  88.2& 80.4& 76.4& 73.2& 70.0& 67.2& 64.5& 62.7& 61.5 & 71.6 & 26.7 \\
    %{\textbf{ILAR}} & 71.32 & 67.477 & 63.486 & 59.493 & 56.425 & 53.341 & 51.2 & 49.326 & 47.53 & \textbf{23.79}\\
    \hline
    \hline
    %CLIP ZSL* (ResNet101) & 81.8 & 81.1 & 79.6 & 79.8 & 79.0 & 78.8 & 79.3 & 79.9 & 80.1 & 79.9 & 1.7\\
    CLIP ZSL* & ViT-B/16 & 85.8 & 85.5 & 84.9 & 84.8 & 84.2 & 84.2 & 84.0 & 83.9 & 83.7 & 84.6 & 2.1 \\
    %AttriCLIP]
    AttriCLIP \cite{wang2023attriclip} & ViT-B/16 & 90.3  & 76.4 & 67.7 & 60.2 & 53.9 & 48.3 & 44.0 & 41.1 & 37.8 & 57.7 & 52.5 \\
    LGCL \cite{khan2023introducing} & ViT-B/16 & 95.2 & 81.6 & 71.4 & 63.5 & 57.4 & 51.93 & 47.6 & 43.94 & 40.8 & 61.5 & 54.4 \\
    SV-T \cite{qiu2023semantic} &ViT-B/16 & 90.6 & 89.2 & 86.8 & 85.4 & 84.8 & 83.4 & 81.9 & 81.9 & 81.7 & 85.1 & 8.9\\
    %CPE & 90.23 & 89.56 & 87.42 & 86.80 & 86.51 & 85.08 & 83.43 & 83.38 & 82.77 \\
    CoCoOp \cite{zhou2022conditional}* &ViT-B/16 & 94.2& 91.7& 88.9& 87.8& 86.3& 84.3& 82.5& 81.9& 81.3 & 86.5 & 12.9 \\
    \hline
    %\textbf{Ours} (ResNet101) & 5 & \cmark & \cmark & MLP1  & 87.1 & 81.5 & 79.6 & 78.1 & 76.8 & 74.3 & 74.2 & 74.1 & 73.3 & 77.7 & 13.8\\     
    %\textbf{Ours} (ResNet101) & 91.2 & 87.3 & 86.4 & 86.2 & 85.4 & 85.1 & 85.0 & 84.3 & 84.0 & 86.1 & 7.2\\     
    %\textbf{Ours} (ViT-B/16) &  95.6 & 93.7 & 90.0 & 86.7 & 85.4 & 83.7 & 84.2 & 83.9 & 83.1 & 87.4 & 12.5 \\ 
    %\textbf{Ours} & 5 & \cmark & \cmark & MLP2  & \\ 
    %\textbf{Ours} (ViT-B/16)& 5 & \xmark & \cmark & MLP1  & 91.3 & 85.2 & 81.3 & 76.2 & 73.4 & 68.4 & 62.3 & 59.3 & 56.2 & 72.6 & 35.1 \\ 
    %\textbf{Ours} & 5 &  & & 95.5 & 94.1 & 91.6 & 89.8 & 88.6 & 87.5 & 87.4 & 86.9 & 86.4 \\ 
    %\textbf{Ours} (ViT-B/16)& 5 & \cmark & \xmark & MLP1  & 21.5 & 5.3 & 4.3 & 3.8 & 4.0 & 3.9 & 3.7 & 3.6 & 3.6 & 5.9 & 17.1  \\
    %\textbf{Ours} (ViT-B/16)& 5 & \cmark & \cmark & MLP2  & 95.5 & 94.1 & 91.6 & 91.2 & 89.5 & 88.9 & 89.2 & 88.6 & 88.3 & 90.8 & 7.2\\ 
    %\textbf{Ours} (ViT-B/16)& 5 & \cmark & \cmark & RES2  & 95.5 & 94.4 & 93.3 & 92.2 & 91.1 & 90.6 & 90.6 & 90.4 & 89.9 & 92.0 & 5.6\\ 
    %\textbf{Ours} (ResNet50x4) & 5 & \cmark & \cmark & MLP2  & \\     
    \textbf{Ours} &ViT-B/16 & 95.4 & 94.4 & 93.4 & 93.1 & 92.1 & 91.4 & 90.8 & 90.0 & \textbf{\underline{89.1}} & \textbf{\underline{92.2}} & 6.3\\ 
    \thickhline
    \end{tabular}
    \end{adjustbox}
    \label{table:sota_miniimagenet}
\caption{Comparison with state-of-the-art methods on the miniImageNet dataset with 5-way 5-shot settings. $\uparrow$ means higher is better, while $\downarrow$ denotes lower is better. The double horizontal line distinguish whether the CLIP model was used. 
* mark denotes the re-implemented results, where we re-implemented the CLIP ZSL (zero-shot learning) and CoCoOp for the FSCIL task. Also, we re-implemented ALICE and BSC using ViT-B backbone. 
The values for SAVC \cite{song2023learning}\dag are estimated values from the figure.
}
\label{app:table:sota_mini}
\end{table*}

\begin{table*}[!t]
\centering
\sisetup{table-format=4.0} % integer values only, up to 4 digits
    \renewcommand{\arraystretch}{1.1}
    \centering
    \begin{adjustbox}{max width=1.0\textwidth}
    %\begin{adjustbox}{angle=90}
    \begin{tabular}{l l l l c c c c c c c c c  l}
    \thickhline
    \multirow{2}{*}{\textbf{Method}} &  \multirow{2}{*}{\shortstack{Backbone}} & \multicolumn{9}{c}{\textbf{Acc. in each session$\uparrow$ (\%) }}& \multirow{2}{*}{\textbf{AVG($\uparrow$)}} & \multirow{2}{*}{\textbf{PD($\downarrow$)}} \\ 
    %\cline{2-10}
    \cline{3-11}
    & & 1 & 2 & 3 & 4 & 5 & 6 & 7 & 8 & 9 \\
    \hline
    TOPIC \cite{tao2020few} & ResNet18& 64.1 & 55.9 & 47.1 & 45.2 & 40.1 & 36.4 & 34.0 & 31.6 & 29.4 & 42.6 & 34.7 \\
    %\hline
    ERL++ \cite{dong2021few} & ResNet18& 73.6 & 68.2 & 65.1 & 61.8 & 58.4 & 55.5 & 52.5 & 50.2 & 48.2 & 59.3& 25.4 \\
    CEC \cite{zhang2021few} & ResNet18& 73.1 & 68.9 & 65.3 & 61.2 & 58.1 & 55.6 & 53.2 & 51.3 & 49.1 & 59.5 & 23.9\\
    %\hline
    Data replay \cite{liu2022few} & ResNet18& 74.4 & 70.2 & 66.5 & 62.5 & 59.7 & 56.6 & 54.5 & 52.4 & 50.1 & 60.8 & 24.3\\
    MCNet \cite{ji2022memorizing} & ResNet18& 73.3 & 69.3 & 65.7 & 61.7 & 58.8 & 56.4 & 54.6 & 53.0 & 50.7 & 60.4 & 22.6 \\
    LIMIT \cite{zhou2022few} & ResNet18 & 73.8 & 72.1 & 67.9 & 63.9 & 60.7 & 57.8 & 55.7 & 53.5 & 51.2 & 61.8 & 22.6 \\
    FACT \cite{zhou2022forward} & ResNet18 & 74.6 & 72.1 & 67.6 & 63.5 & 61.4 & 58.4 & 56.3 & 54.2 & 52.1 &62.2 & 22.5 \\
    CLOM \cite{zou2022margin} & ResNet18& 74.2 & 69.8 & 66.2 & 62.4 & 59.3 & 56.5 & 54.4 & 52.2 & 50.3 & 60.6 & 24.0 \\
    C-FSCIL\cite{hersche2022constrained}& ResNet18 & 77.5 & 72.4 & 67.5 & 63.3 & 59.8 & 57.0 & 54.4 & 52.5 & 50.5 & 61.6 & 27.0 \\ 
    %\hline
    ALICE \cite{peng2022few}& ResNet18 & 79.0 & 70.5 & 67.1 & 63.4 & 61.2 & 59.2 & 58.1 & 56.3 & 54.1 & 63.2 & 24.9 \\
    SAVC \cite{song2023learning}\dag & ResNet18 & 79.0 & 73.0 & 70.0 & 65.0 & 62.0 & 59.0 & 57.0 & 55.0 & 52.0 & 63.6 & 27.0\\ 
    BSC \cite{yoon2023balanced} & ResNet18 &  74.3 & 70.5 & 68.5 & 65.1 & 62.4 & 60.4 & 58.6 & 56.4 & 54.7 & 63.4 & 19.6 \\
    ALICE \cite{peng2022few} & ViT-B & 85.3 & 80.7 & 76.1 & 71.5 & 68.1 & 65.9 & 62.1 & 60.1 & 57.3 & 69.7 & 28.0 \\
    BSC \cite{yoon2023balanced} & ViT-B & 85.7 & 81.5 & 76.3 & 72.2 & 69.4 & 66.1 & 64.9 & 62.5 & 58.8 & 70.8 & 26.9 \\
    %{\textbf{ILAR}} & 71.93 & \textbf{67.29} & \textbf{63.07} & \textbf{59.8} & \textbf{56.96} & \textbf{54.54} & \textbf{52.06} & \textbf{49.97} & \textbf{47.92} & 24.01\\
    \hline
    \hline
    CLIP ZSL* & ViT-B/16 & 73.8 & 71.6 & 72.0 & 70.71 & 69.8 & 68.5 & 67.8 & 67.3 & 66.9 & 69.8 & 6.9\\
    SV-T \cite{qiu2023semantic} & ViT-B/16 & 86.8 & 82.8 &  80.4 & 77.2 & 76.1 & 74.0 & 72.9 & 71.7 &  69.8 &76.9 & 17.0 \\
    %CPE & 87.83 & 85.86 & 84.93 & 82.85 & 82.64 & 82.42 & 82.27 & 81.44 & 80.52 \\
    %AttriCLIP
    CoCoOp \cite{zhou2022conditional}* &ViT-B/16 & 82.2 & 77.4 & 73.7 & 71.7 & 69.1 & 67.7 & 65.5 & 63.8 & 62.1 & 70.4 & 20.1 \\
    \hline
    \textbf{Ours} & ViT-B/16 &  86.2 & 82.3 & 79.6 & 77.5 & 75.9 & 75.3 & 74.1 & 73.6 & \textbf{\underline{72.8}} & \textbf{\underline{77.5}} & 13.4 \\
    \thickhline
    \end{tabular}
    \end{adjustbox}
    \label{table:sota_cifar100}
\caption{Comparison with state-of-the-art methods on the CIFAR100 with 5-way 5-shot setting. $\uparrow$ means higher is better, while $\downarrow$ denotes lower is better. The double horizontal line distinguish whether the CLIP model was used. 
* mark denotes the re-implemented results, where we re-implemented the CLIP ZSL (zero-shot learning) and CoCoOp for the FSCIL task. Also, we re-implemented ALICE and BSC using ViT-B backbone. 
The values for SAVC \cite{song2023learning}\dag are estimated values from the figure.
%, as larger CLIP models inherently exhibit superior performance.%; hence, consistent use of the same model type is necessary for fair comparison. 
%The 'Coreset' column indicates the number of features saved for each class, 'cls-wise' specifies whether cls-wise token embedding was used, 'Init.' denotes whether the initialization trick was applied, and 'Key-map' refers to the model type for the key map.
%Comparison with the state-of-the-art methods on miniImageNet dataset with 5-way 5-shot setting. $\uparrow$ means the higher is the better, while $\downarrow$ denotes the lower is the better. Horizontal double line distinguish whether the CLIP model was used. The result of CLIP ZSL (zero-shot learning) and CoCoOp is re-implemented for the FSCIL task. Note that we only included the results that clearly clarified the type of CLIP model used, since the bigger CLIP model necessarily shows better performance so consistent use of the model type is needed for the fair comparison. Column `Coreset denotes the number of feature saved for each class, `cls-wise' denotes whether cls-wise token embedding was used, `Init.' denotes whether the initialization trick was used, and `Key-map' denotes the model type for key map.
%Unreported results with no codes available are excluded from the table. 
}
\label{app:table:sota_cifar}
\end{table*}

\begin{table*}[!t]
\centering
\sisetup{table-format=-1.2}   % 2 decimals, leave space for minus sign
    \renewcommand{\arraystretch}{1.1}
    \centering
    \begin{adjustbox}{max width=1.0\textwidth}
    %\begin{adjustbox}{angle=90}
    \begin{tabular}{l l c c c c c c c c c c c l l }
    \thickhline
    \multirow{2}{*}{\textbf{Method}} &  \multirow{2}{*}{\shortstack{Backbone}} & \multicolumn{11}{c}{\textbf{Acc. in each session$\uparrow$ (\%) }} & \multirow{2}{*}{\textbf{AVG($\uparrow$)}} & \multirow{2}{*}{\textbf{PD($\downarrow$)}} \\ 
    \cline{3-13}
    & & 1 & 2 & 3 & 4 & 5 & 6 & 7 & 8 & 9 & 10 & 11 \\
    \hline
    TOPIC \cite{tao2020few} & ResNet18& 68.7 & 62.5 & 54.8 & 50.0 & 45.3 & 41.4 & 38.4 & 35.4 & 32.2 & 28.3 & 26.3 & 43.9 & 42.4 \\
    IDLVQ-C \cite{chen2020incremental} & ResNet18& 77.4 & 74.7 & 70.3 & 67.1 & 65.3 & 63.5 & 62.1 & 61.5 & 59.0 & 58.7 & 57.8 & 65.2 & 19.6 \\
    SKD \cite{cheraghian2021semantic} & ResNet18& 68.2 & 60.5 & 55.7 & 50.5 & 45.7 & 42.9 & 40.9 & 38.8 & 36.5 & 34.8 & 33.0 & 46.1 &  35.3
     \\
     ERL++ \cite{dong2021few} & ResNet18& 73.5 & 71.1 & 66.1 & 63.3 & 59.5 & 59.9 & 58.6 & 57.7 & 56.2 & 54.8 & 52.3 & 61.2 &  21.2 \\
     CEC \cite{zhang2021few} & ResNet18& 75.9 & 71.9 & 68.5 & 63.5 & 62.4 & 58.3 & 57.7 & 55.8 & 54.8 & 53.5 & 52.3 & 61.3 & 23.6\\
    Data replay \cite{liu2022few} & ResNet18& 75.9 & 72.1 & 68.6 & 63.8 & 62.6 & 59.1 & 57.8 & 55.9 & 54.9 & 53.6 & 52.4 & 61.5 & 23.5 \\
    MCNet \cite{ji2022memorizing} & ResNet18&  77.6 & 74.0 & 70.5 & 65.8 & 66.2 & 63.8 & 62.1 & 61.8 & 60.4 & 60.1 & 59.1 & 65.6 & 18.5 \\
    LIMIT \cite{zhou2022few} & ResNet18& 75.9 & 73.6 & 72.0 & 68.1 & 67.4 & 63.6 & 62.4 & 61.4 & 59.9 & 58.7 & 57.4 & 65.5 & 18.5 \\
    FACT \cite{zhou2022forward}& ResNet18 & 75.9 & 73.2 & 70.8 & 66.1 & 65.6 & 62.2 & 61.7 & 59.8 & 58.4 & 57.9 & 56.9 & 64.4& 19.0 \\
    CLOM \cite{zou2022margin}& ResNet18 & 79.6 & 76.1 & 72.9 & 69.8 & 67.8 & 65.6 & 63.9 & 62.6 & 60.6 & 60.3 & 59.6 & 67.2 & 20.0 \\
    ALICE \cite{peng2022few} & ResNet18& 77.4 & 72.7 & 70.6 & 67.2 & 65.9 & 63.4 & 62.9 & 61.9 & 60.5 & 60.6 & 60.1 & 65.7 & 17.3 \\
    SAVC \cite{song2023learning}\dag & ResNet18 &   81.9 & 77.9 & 75.0 & 70.2 & 70.0 & 67.0 & 66.2 & 65.3 & 63.8 & 63.2 & 62.5 & 69.4 & 19.4\\     
    BSC \cite{yoon2023balanced}& ResNet18& 80.1 & 76.6& 74.0 & 72.0 & 70.41 & 70.3 & 69.2 & 66.3 & 65.6 & 64.4 & 63.0 & 70.2 &  17.1 \\
    \hline
    \hline
    CLIP ZSL* & ViT-B/16 & 65.5 & 64.2 & 63.2 & 62.4 & 59.9 & 60.3 & 59.8 & 58.4 & 56.3 & 54.9 & 53.5 & 59.9 & 12.0 \\
    %SV-T \cite{qiu2023semantic} & ViT-B/16\\
    CoCoOp \cite{zhou2022conditional}*& ViT-B/16 & 80.3 & 72.1 & 68.8 & 65.4 & 63.4 & 61.2 & 58.2 & 56.9 & 54.5 & 52.3 & 50.1 & 62.1 & 30.2\\
    \hline
    \textbf{Ours} & ViT-B/16 & 81.3& 77.4& 75.8& 73.3 & 72.6& 70.4& 68.7& 67.3& 65.9& 64.4& 63.8 &71.0 & 17.5\\   
    %{\textbf{Ours}} & 76.92 & 71.95 & 70.18 & 66.94 & 63.99 & 61.38 & 59.47 & 57.29 & 55.43 & 54.98 & \underline{53.16} & 23.76 \\
    \thickhline
    \end{tabular}
    \end{adjustbox}
    \label{app:table:sota_cub200_pretrain}
    %\end{table*}
\caption{Comparison with state-of-the-art methods on the CUB200 dataset with 10-way 5-shot setting. $\uparrow$ means higher is better, while $\downarrow$ denotes lower is better. The double horizontal line distinguish whether the CLIP model was used. 
* mark denotes the re-implemented results, where we re-implemented the CLIP ZSL (zero-shot learning) and CoCoOp for the FSCIL task. Also, we re-implemented ALICE and BSC using ViT-B backbone. 
Note that for the CUB200 dataset, the previous methods without using the CLIP model utilize ImageNet-pretrained model.
%, as larger CLIP models inherently exhibit superior performance.%; hence, consistent use of the same model type is necessary for fair comparison. 
%The 'Coreset' column indicates the number of features saved for each class, 'cls-wise' specifies whether cls-wise token embedding was used, 'Init.' denotes whether the initialization trick was applied, and 'Key-map' refers to the model type for the key map.
%Comparison with the state-of-the-art methods on miniImageNet dataset with 5-way 5-shot setting. $\uparrow$ means the higher is the better, while $\downarrow$ denotes the lower is the better. Horizontal double line distinguish whether the CLIP model was used. The result of CLIP ZSL (zero-shot learning) and CoCoOp is re-implemented for the FSCIL task. Note that we only included the results that clearly clarified the type of CLIP model used, since the bigger CLIP model necessarily shows better performance so consistent use of the model type is needed for the fair comparison. Column `Coreset denotes the number of feature saved for each class, `cls-wise' denotes whether cls-wise token embedding was used, `Init.' denotes whether the initialization trick was used, and `Key-map' denotes the model type for key map.
%Unreported results with no codes available are excluded from the table. 
}
\label{app:table:sota_cub}
\end{table*}


\end{document}